\documentclass{article}
\pdfoutput=1

\usepackage{microtype}
\usepackage{graphicx}
\usepackage{booktabs} % for professional tables
\usepackage{tcolorbox}
\usepackage{caption} 

\usepackage{hyperref}

\usepackage[accepted]{icml2025}

\usepackage{amsmath}
\usepackage{amssymb}
\usepackage{mathtools}
\usepackage{amsthm}

\usepackage[capitalize,noabbrev]{cleveref}

\usepackage{algorithm}
\usepackage{algpseudocode}

\usepackage{enumitem}
  \newlist{inlinelist}{enumerate*}{1}
  \setlist*[inlinelist,1]{%
          label=(\roman*),
      }

\usepackage{adjustbox}

\theoremstyle{plain}

\theoremstyle{definition}

\theoremstyle{remark}

\usepackage[textsize=tiny]{todonotes}

\icmltitlerunning{JoLT: Joint Probabilistic Predictions on Tabular Data Using LLMs}

\begin{document}

\twocolumn[
\icmltitle{JoLT: Joint Probabilistic Predictions on Tabular Data Using LLMs}



\icmlsetsymbol{equal}{*}

\begin{icmlauthorlist}
\icmlauthor{Aliaksandra Shysheya}{equal,cam}
\icmlauthor{John Bronskill}{equal,cam}
\icmlauthor{James Requiema}{tor,vec}
\icmlauthor{Shoaib Ahmed Siddiqui}{cam}
\icmlauthor{Javier González}{msft}
\icmlauthor{David Duvenaud}{tor,vec}
\icmlauthor{Richard E. Turner}{cam,ati}
\end{icmlauthorlist}

\icmlaffiliation{cam}{University of Cambridge}
\icmlaffiliation{tor}{University of Toronto}
\icmlaffiliation{ati}{The Alan Turing Institute}
\icmlaffiliation{vec}{Vector Institute}
\icmlaffiliation{msft}{Microsoft Research Cambridge}

\icmlcorrespondingauthor{John Bronskill}{jfb54@cam.ac.uk}


\vskip 0.3in
]



\printAffiliationsAndNotice{\icmlEqualContribution} % otherwise use the standard text.

\begin{abstract}
We introduce a simple method for probabilistic predictions on tabular data based on Large Language Models (LLMs) called JoLT (Joint LLM Process for Tabular data). 
JoLT uses the in-context learning capabilities of LLMs to define joint distributions over tabular data conditioned on user-specified side information about the problem, exploiting the vast repository of latent problem-relevant knowledge encoded in LLMs.  
JoLT defines joint distributions for multiple target variables with potentially heterogeneous data types without any data conversion, data preprocessing, special handling of missing data, or model training, making it accessible and efficient for practitioners. 
Our experiments show that JoLT outperforms competitive methods on low-shot single-target and multi-target tabular classification and regression tasks.
Furthermore, we show that JoLT can automatically handle missing data and perform data imputation by leveraging textual side information.
We argue that due to its simplicity and generality, JoLT is an effective approach for a wide variety of real prediction problems.
\end{abstract}
\section{Introduction}
\label{sec:introduction}
Tabular data is ubiquitous in machine learning applications in finance, medicine, business, agriculture, and education \citep{sahakyan2021explainable,fang2024large}. Given their ubiquitous nature, \citet{van2024position} argue that tabular foundation models should rank higher as a research priority.
However, real-world tabular data rarely presents itself in an idealized form -- practitioners often encounter datasets with mixed data types, missing values, and complex dependencies between variables.
Traditional approaches typically require extensive preprocessing pipelines and specialized handling for different data types. %creating barriers to deployment. 
\definecolor{OliveGreen}{RGB}{34,139,34}

\begin{table*}[h]
 % \caption{Comparison of JoLT key attributes with competitive methods.}
\label{tab:comparison}
\vskip -0.1in
\centering
\begin{small}
\caption{\textbf{JoLT: key features and comparison with competitive methods.} JoLT automatically handles missing data, supports mixed data types, provides joint probabilities by default, requires no training, and effectively leverages side information. However, JoLT is slower than some competitive methods and does not easily scale to large tables.}
\begin{adjustbox}{max width=\textwidth}
\begin{tabular}{l c c c c c c c}
\toprule
\text{Method} & \begin{tabular}[c]{@{}c@{}}\text{Automatically handle}\\ \text{missing data}\end{tabular} & \begin{tabular}[c]{@{}c@{}}\text{Mixed types}\\ \text{+ strings}\end{tabular} & \begin{tabular}[c]{@{}c@{}}\text{Can use}\\ \text{side info}\end{tabular} & \begin{tabular}[c]{@{}c@{}}\text{In-context learning}\\ \text{(no training)}\end{tabular} & \begin{tabular}[c]{@{}c@{}}\text{Joint probabilities} \\ \text{by default}\end{tabular} & \begin{tabular}[c]{@{}c@{}}\text{Fast at} \\ \text{test time}\end{tabular} & \begin{tabular}[c]{@{}c@{}}\text{Scalable to} \\ \text{large tables}\end{tabular} \\
\midrule
\text{TabLLM} & \textcolor{red}{$\times$} & \textcolor{OliveGreen}{$\checkmark$} & \textcolor{OliveGreen}{$\checkmark$} & \textcolor{red}{$\times$} & \textcolor{red}{$\times$}  & \textcolor{red}{$\times$} & \textcolor{red}{$\times$}\\
\text{TabPFN} & \textcolor{red}{$\times$} & \textcolor{red}{$\times$} & \textcolor{red}{$\times$} & \textcolor{OliveGreen}{$\checkmark$} & \textcolor{red}{$\times$} & \textcolor{OliveGreen}{$\checkmark$} & \textcolor{red}{$\times$} \\
\text{XGBoost} & \textcolor{red}{$\times$} & \textcolor{red}{$\times$} & \textcolor{red}{$\times$} & \textcolor{red}{$\times$} & \textcolor{red}{$\times$} & \textcolor{OliveGreen}{$\checkmark$} & \textcolor{OliveGreen}{$\checkmark$} \\
\midrule
\text{JoLT (Ours)} & \textcolor{OliveGreen}{$\checkmark$} & \textcolor{OliveGreen}{$\checkmark$} & \textcolor{OliveGreen}{$\checkmark$} & \textcolor{OliveGreen}{$\checkmark$} & \textcolor{OliveGreen}{$\checkmark$} & \textcolor{red}{$\times$} & \textcolor{red}{$\times$} \\
\bottomrule
\end{tabular}
\end{adjustbox}
\end{small}
\end{table*}
When modeling tabular data, incorporating side information and domain expertise is often a challenge. Additionally, many current approaches are inaccessible to lay users who lack advanced technical knowledge. 
For this reason, using Large Language Models (LLMs) to incorporate textual information for tabular data prediction \citep{fang2024large, lu2024large} is a natural approach. LLMs, trained on vast corpora of internet data, possess substantial implicit knowledge and provide an interface for expressing and incorporating side information in natural language. These attributes make LLMs particularly attractive for tabular prediction, offering the ability to integrate diverse data types naturally while circumventing the need for complex preprocessing pipelines.

Prediction using LLMs typically uses one of two approaches -- fine-tuning \citep{yosinski2014transferable} or inference-time in-context learning (ICL) \citep{brown2020language}. 
Fine-tuning involves modifying LLM's weights to adapt it for specific tasks. This approach has its own limitations, including high resource requirements, overfitting risks with small datasets, and potential privacy concerns when using sensitive data. 
In contrast, ICL is a simpler, more accessible alternative, enabling predictions without gradient updates to the model. 
This approach reduces the burden on users by eliminating the need for training or hyperparameter tuning, making it suitable for scenarios with limited data, such as personalization~\citep{massiceti2021orbit, ding2017collecting}.
\begin{figure*}[t]
% \vskip 0.2in
\begin{center}
\centerline{\includegraphics[width=0.9\textwidth]{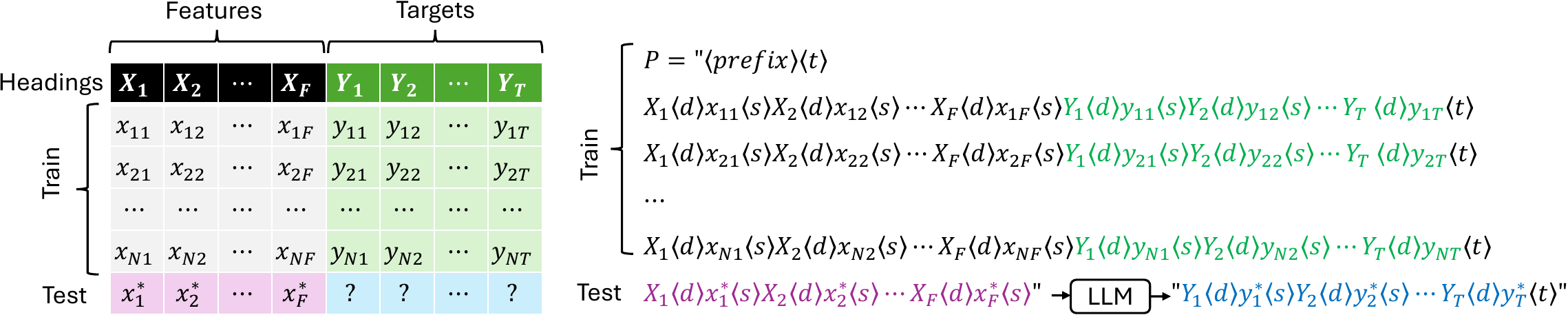}}
\vspace{-3mm}
\caption{\textbf{Mapping tabular data to a prompt $P$.} The diagram on the left depicts a tabular dataset where the first $N$ rows are training examples with $F$ features and $T$ targets fully observed. The last row represents a test example with $F$ observed features and unobserved targets. The prompt at the top right is formed by serializing the $N$ training examples and the test features into a single string. The prompt is input to a pretrained LLM that will generate the targets to complete the test example. See \cref{tab:nomenclature} for nomenclature.}
\label{fig:table_to_prompt}
\end{center}
\vspace{-8mm}
\end{figure*}

Well-calibrated uncertainty quantification is critical for decision-making processes that rely on tabular predictions \citep{sahakyan2021explainable,bishop2006pattern,kochenderfer2015decision}. 
Additionally, joint probabilistic predictions have practical applications in sciences~\citep{kong2020deep} and healthcare, where they help predict multiple patient outcomes, such as survival, recovery time, and complication risk~\citep{martin2021clinical}.
Joint modeling can also exploit correlations between outputs to improve prediction accuracy \citep{MorenoArtesAlvarez18}.

\citet{requeima2024llm} introduced LLM Processes (LLMPs) which used pretrained LLMs and ICL to do probabilistic regression conditioned on textual side information. LLMPs effectively combine data and metadata, such as dataset descriptions and column headings, with the LLM’s latent knowledge. 
However, LLMPs are not currently suitable for tabular data as they do not support heterogeneous columns, including numerical and categorical types, or missing data, which many existing methods also fail to address comprehensively.
In this work, we build upon LLMPs and present JoLT (Joint LLMP for Tabular data), a method designed to address the challenges of heterogeneous tabular data. JoLT extends LLMPs beyond regression to make joint probabilistic predictions for multiple target variables, accommodating both numerical and categorical data types. 
Our approach eliminates the need for preprocessing, missing data imputation, model training, or hyperparameter tuning, making it highly accessible to non-experts in machine learning. 
Additionally, JoLT empowers users to incorporate side information or provide specific instructions in plain language or text, allowing for seamless integration of expert knowledge and contextual insights from the user and LLM into the modeling process. 
Our main contributions are:
\vspace{-3mm}
\begin{itemize}[leftmargin=*]
\setlength\itemsep{0pt}
    \item We present JoLT, a method that extends LLMPs beyond regression to make joint probabilistic predictions for multiple target variables with heterogeneous data types on tabular datasets.
    \item We demonstrate that JoLT outperforms competitive approaches on low-shot single and multiple target tabular classification and regression tasks.
    \item We show that JoLT implicitly handles missing tabular data and performs as well or better on downstream tasks when compared to imputing data as a preprocessing step.
    \item Finally, we show that JoLT can also effectively impute missing data by leveraging textual side information about the problem.
\end{itemize}
\vspace{-4mm}
Table \labelcref{tab:comparison} summarizes the key attributes of JoLT relative to competitive methods.
\section{Method}
\label{sec:method}
In this section, we describe how to make heterogeneous multiple target probabilistic predictions in the presence of missing data using JoLT.

\subsection{Prompt Engineering}
In a prediction setting, tabular datasets consist of multiple rows of examples or records.
Each row consists of one or more columns of features and one or more columns of targets to be predicted based on the feature values. 
Training examples have both observed features and observed targets, while test examples only have observed features.
\begin{table}[t]
\vspace{-2mm}
  \centering
  \caption{Nomenclature used in prompt construction.}
  \label{tab:nomenclature}%
  % \vskip 0.15in
  \begin{small}
  % \begin{sc}
  \begin{adjustbox}{max width=\columnwidth}
    \begin{tabular}{cl}
    \toprule
    \multicolumn{1}{c}{\textbf{Symbol}} & \multicolumn{1}{l}{\textbf{Description}} \\
    \midrule
     $P$ & prompt\\
     $F$ & number of features\\
     $T$ & number of targets\\
     $N$ & number of training examples or shots\\
     $M$ & number of text examples\\
     ${X_1, X_2, \dots, X_F}$ & text heading for feature columns\\
     ${Y_1, Y_2, \dots, Y_T}$ & text heading for target columns\\
     $x_{i,j}$ &  $j$th feature value of the $i$th training example\\
     $y_{i,j}$ &  $j$th target value of the $i$th training example\\
     $x^*_{i,j}$ &  $j$th feature value of the $i$th test example\\
     $y^*_{i,j}$ &  $j$th target value of the $i$th test example\\
     $\langle prefix \rangle$ & text string with side information\\
     $\langle d \rangle$ & separates $X_j$ and $x_{i,j}$ or $Y_j$ and $y_{i,j}$\\
     $\langle s \rangle$ & separates $X_j\langle d \rangle x_{i,j}$ and $X_{j+1}\langle d \rangle x_{i,j+1}$\\
      & or $Y_j\langle d \rangle y_{i,j}$ and $Y_{j+1}\langle d \rangle y_{i,j+1}$\\
     $\langle t \rangle$ & separates examples\\
    \bottomrule
    \end{tabular}%
  \end{adjustbox}
  % \end{sc}
  \end{small}
\vspace{-3mm}
\end{table}%

\cref{fig:table_to_prompt} shows how we design a prompt that contains the entire training set and the features for a test example that is fed to the LLM in order to generate a prediction.
For each test example, we serialize each row of training data, including both features and targets, plus the features of the test example to a string.
Each feature column may be of any data type (e.g. numerical - integer or floating point, categorical, date/time, addresses, unstructured text, etc.) given that the type is or can be converted to a string.
The target columns can also be of any type. However, we currently only support computing predictive distributions for numerical and categorical types.
We do not perform any preprocessing or scaling of the data as we want to retain the natural scale and units of the data such that related knowledge encoded in the LLM can be leveraged.
Furthermore, we do not modify the LLM weights via training or fine-tuning.
Next, we discuss how to make predictions and compute joint predictive distributions based on the prompt.
\subsection{Prediction}
Here we present two approaches to making predictions on multiple target, heterogeneous data -- rejection sampling and sampling from a full distribution.

\textbf{Rejection Sampling}
After feeding the prompt to the LLM, we use the autoregressive token prediction capability of the LLM to generate the targets for the test example.
Predicting most targets will require multiple steps of autoregressive token generation to represent numerical digits or categories.
Knowing the number and types of the targets, we can parse the generated output and ensure it conforms to the expected format.
If not, we reject the sample.
For numerical targets, we ensure the sample contains a valid number, and for categorical targets, we ensure that it contains a known category.
For a point estimate of a quantity, we can take a single top 1 sample from the LLM.
To obtain a point estimate and uncertainty for numerical targets, we can take a set of samples and use the median for the point estimate and compute a confidence interval over the range.
For categorical targets, the set of samples form a categorical distribution and the point estimate is the category with the highest number of samples.
To illustrate, \cref{fig:uncertainty} depicts median point estimates and the 95$\%$ confidence interval for JoLT predictions of alcohol percentage on the wine quality dataset \citep{wine_quality_186}.
In this example, the predictive medians are quite accurate and the confidence intervals well-calibrated.

\textbf{Full Distribution via LLM Logits}
We can compute the probability of a target value $s$ that is a member of the set of possible target values $\mathcal{S}$ conditioned on a prompt $P$ as:
\begin{equation}
\label{eq:target_probability}
    p(y = s | P, s \in S) = \frac{p(y = s | P)}{\sum_{s^\prime \in \mathcal{S}} p(y=s^\prime | P)}
\end{equation}
For categorical targets, we can obtain $p(y = s | P)$ for each category $s$ from the LLM logits using \cref{alg:categorical_logprobs} and then use \cref{eq:target_probability} to get the normalized probability for each category.
We can then predict the target category that has the highest probability or sample from the resulting full categorical distribution.
If the number of classes is relatively small, this approach is preferable to rejection sampling, as it enables direct access to the full predictive distribution.
However, for numerical targets or when the number of classes is large, the rejection sampling approach is preferred over the logits method. 
This is because the set $\mathcal{S}$ becomes large, and computing the full predictive distribution would require an excessive number of calls to the LLM to retrieve logits, making the logits method computationally impractical.
\subsection{Computing Joint Predictive Distributions}
Using the product rule, we can compute the joint probability of the ground truth target values for any test example as:
\begin{multline}
\label{eqn:joint_probability}
    p(y^*_1, y^*_2, \dots, y^*_T) =
    p(y^*_1 | ``P Y_1 \langle d \rangle")\\
    p(y^*_2 | ``P Y_1 \langle d \rangle y^*_1 \langle s \rangle Y_2 \langle d \rangle") \dots \\
    p(y^*_T | ``P Y_1 \langle d \rangle y^*_1 \langle s \rangle Y_2 \langle d \rangle y^*_2 \langle s \rangle \dots Y_T \langle d \rangle")
\end{multline}
where we have used the notation introduced in \cref{fig:table_to_prompt}. \cref{eqn:joint_probability} suggests that we can compute the joint likelihood of a multiple target test example by computing the product of the probability of each individual target conditioned on the text that precedes it using the logits of the LLM which hold token probabilities.
To introduce dependencies between multiple targets, \cref{eqn:joint_probability} adopts an autoregressive structure. 
This formulation enables the modeling of conditional relationships between the target variables, allowing the prediction of each target to depend on previously predicted ones.
One limitation of this approach is that the predictive distribution becomes dependent on the order of the target variables. As a result, it no longer guarantees a valid stochastic process and may fail to satisfy the Kolmogorov exchangeability condition \citep{oksendal2013stochastic}.

For a numerical target, we follow \citet{requeima2024llm} and approximate the probability of a ground truth value using \cref{alg:numerical_logprobs}.
To obtain a probability density, we convert the predicted probability mass by assuming a uniform distribution within each bin and normalizing by the bin width. 
The bin width is determined by the precision of the generated numerical values.
If the target is categorical, we can use \cref{alg:categorical_logprobs}, with $y$ set to the true target category.
In our experiments, for each test example, we compute the probability for each target, then compute the joint probability using \cref{eqn:joint_probability}, and report the negative log likelihood (NLL) as the mean of the joint NLLs over the test set.
\subsection{Missing Data Handling}
\begin{figure}[t]
% \vskip 0.2in
\begin{center}
\centerline{\includegraphics[width=0.9\columnwidth]{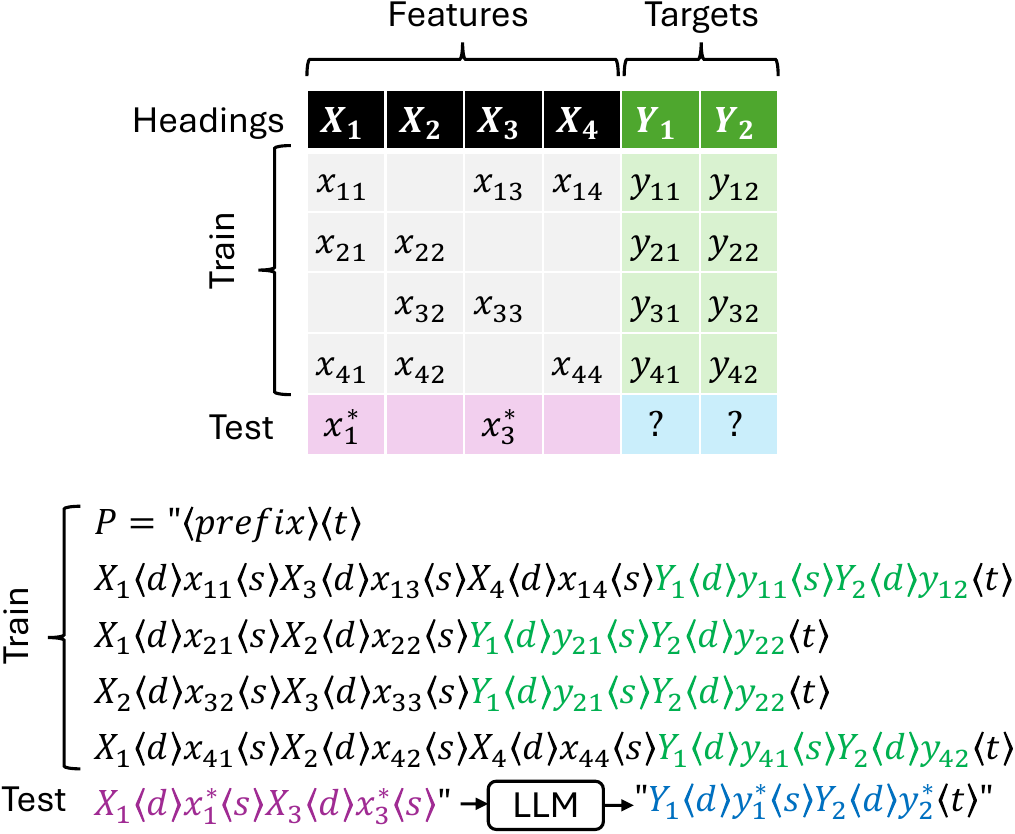}}
\vspace{-3mm}
\caption{\textbf{Missing data handling.} The diagram depicts a tabular dataset with 4 rows of training examples. The last row represents a test example with 2 unobserved targets. Empty cells represent 40$\%$ (8 out of 20 feature cells) missing completely-at-random data. Note that both training and test examples are affected by missing data. The prompt is formed by simply omitting missing cells.}
\label{fig:missing_data_handling}
\end{center}
\vspace{-10mm}
\end{figure}

\label{sec:missing_data}
Our approach implicitly handles missing feature data in both training and test sets.
The strategy is to simply omit any missing feature data when building the prompt as shown in \cref{fig:missing_data_handling}.
In \cref{sec:missing_data_results}, we show that the \textit{omit} approach is either competitive or outperforms missing data imputation.
We posit that formatting the prompt so that each data value is preceded by descriptive column header text indicates to the LLM what data is present versus what data it needs to generate in order to make predictions for the targets.

\subsection{Missing Data Imputation}
\label{sec:imputation}
\begin{figure}[t]
% \vskip 0.2in
\begin{center}
\centerline{\includegraphics[width=1.0\columnwidth]{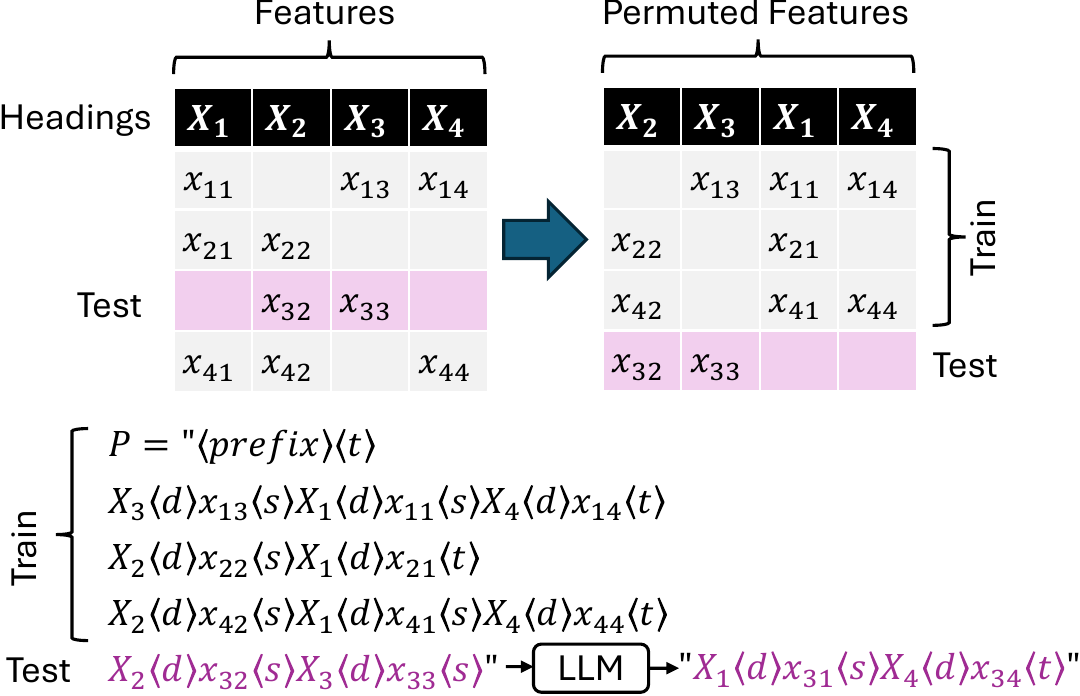}}
\vspace{-3mm}
\caption{\textbf{JoLT imputation.} To impute values for a specific row (highlighted in pink), the features are reordered such that the features with existing values for the specific row are positioned first. The prompt for the LLM is constructed akin to \cref{fig:missing_data_handling}, where instead of predicting targets $Y_i$, the model predicts the missing values for a specific row (e.g., columns $X_1$ and $X_4$ in the diagram). }
\label{fig:imputation_diagram}
\end{center}
\vspace{-10mm}
\end{figure}

Our method can be extended to perform data imputation.
The approach involves imputing missing values row by row, while leveraging other incomplete rows of the table as in-context training data for the LLM. 
To impute missing values for a specific row, the columns of the table are first permuted so that the columns containing missing values for that row are positioned after those with non-missing values.
Following the same procedure described in \cref{sec:missing_data}, a separate prompt is constructed for each row, and the LLM predicts the missing values conditioned on the prompt. 
The procedure is illustrated in \cref{fig:imputation_diagram}.
\section{Experiments}
In this section, we evaluate the performance of JoLT prediction on single- and multiple-target tabular prediction tasks.
Specifically, the experiments aim to answer the following questions:
\begin{inlinelist}
    \item knowing that LLMPs perform well on numerical regression tasks \citep{requeima2024llm}, can JoLT perform well on categorical classification tasks?
    \item how does JoLT compare to strong baseline methods when predicting multiple heterogeneous tabular targets?
    \item how does side information affect JoLT's predictive distribution? 
    \item can JoLT gracefully handle missing data? and
    \item can JoLT impute missing data as well as standard approaches? 
\end{inlinelist}

We use the term \textit{shots} to refer to the number of training examples.
Due to the context size limits on LLMs and the fact that processing time and GPU memory requirement grow roughly quadratically with the length of the prompt, we restrict our experiments to the low-shot setting.
We focus on open-source LLMs, as access to logits for computing probabilities is often unavailable for proprietary LLMs.
In addition, we only use LLMs that perform single-digit tokenization so that we can calculate the probabilities for numerical targets.
As a result, we utilize the Gemma-2 and Qwen2.5 LLMs for all of our experiments.
A key advantage of the JoLT approach is that it can produce probabilistic predictions.
Thus, we restrict the set of competitive approaches to those that can produce a distribution over outputs.
\subsection{Classification Setting}
\label{sec:classification_results}
\begin{figure*}[ht]
%\vskip 0.2in
\begin{center}
\centerline{\includegraphics[width=\textwidth]{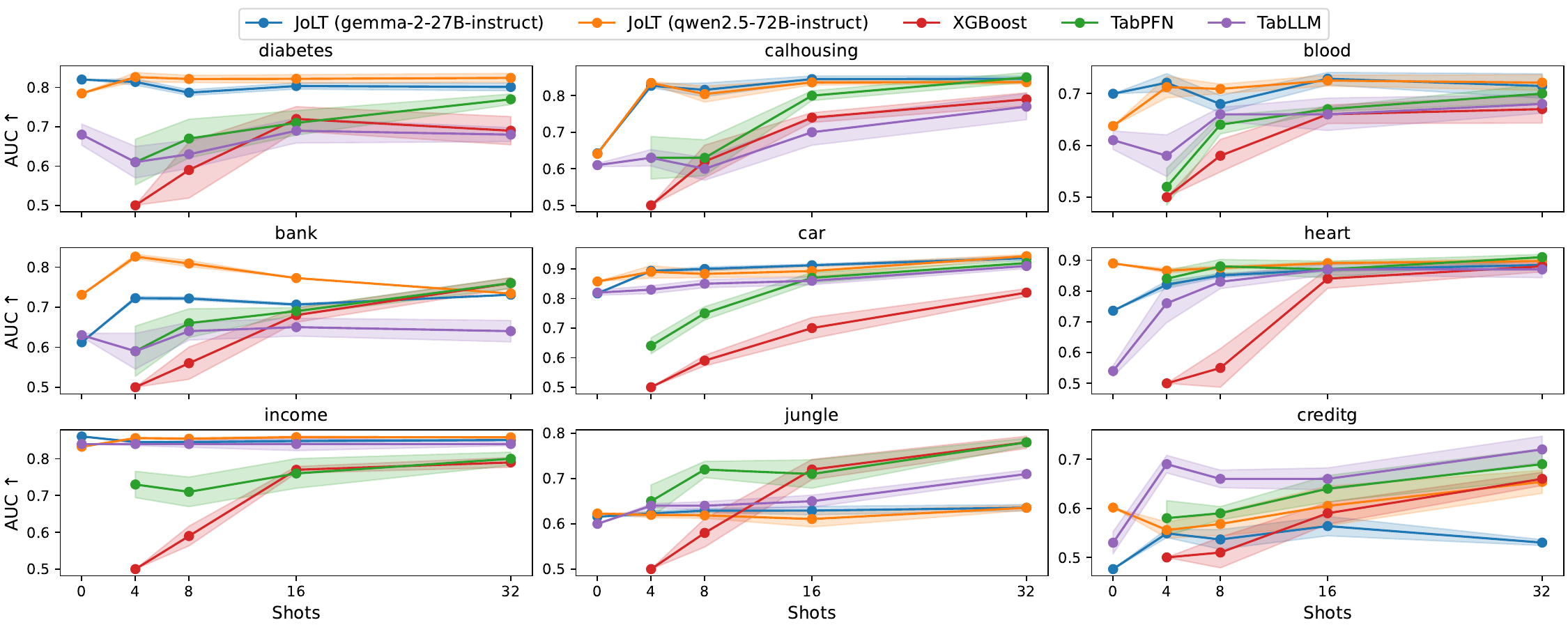}}
\vspace{-4mm}
\caption{Area Under the Receiver Operating Characteristic Curve (AUC) as a function of shot for JoLT using two different LLMs and three competitive methods. The solid line and dots indicate the mean over 5 seeds which affect the training shot selection and the shaded region shows a confidence interval of one $\sigma$. Competitive data from \citet{hegselmann2023tabllm}. Tabular results are in \cref{tab:classification_results}.}
\label{fig:classification_results}
\end{center}
\vspace{-8mm}
\end{figure*}

In this experiment, we compare JoLT low-shot classification performance to strong baselines with the same nine datasets used in \citet{hegselmann2023tabllm}.
To make a fair comparison, we use the serialized datasets from \citet{hegselmann2023github}.
We consider XGBoost \citep{chen2016xgboost}, TabPFN \citep{hollmann2025tabpfn}, and TabLLM \citep{hegselmann2023tabllm} as competitive baselines for comparison.
Like JoLT, TabLLM also relies on an LLM for prediction. However, TabLLM utilizes fine-tuning on the training data as opposed to ICL, which is the main focus of JoLT.

\cref{fig:classification_results} shows that on 7 of the 9 datasets (creditg and jungle being the exceptions), JoLT models outperform other competitive methods in the low-shot setting, often by a large margin.
However, expectedly, the gap shrinks with an increasing number of shots.

\textbf{Summary}: JoLT is able to outperform boosted decision trees (XGBoost), LLM fine-tuning (TabLLM), and TabPFN (which also uses ICL, but without an LLM)  in the low-shot classification setting by utilizing column header side information.
XGBoost and TabPFN cannot use side information and rely on larger amounts of training data to perform well.
TabLLM can leverage text information, but tends to overfit when fine-tuning on a small amount of training data.
\subsection{Multi-target Prediction}
\label{sec:multi_target_results}
Here, we evaluate the ability of JoLT to predict multiple heterogeneous targets and compute joint distributions on three different datasets.
\begin{figure*}[h]
% \vskip 0.2in
\begin{center}
\centerline{\includegraphics[width=1.0\textwidth]{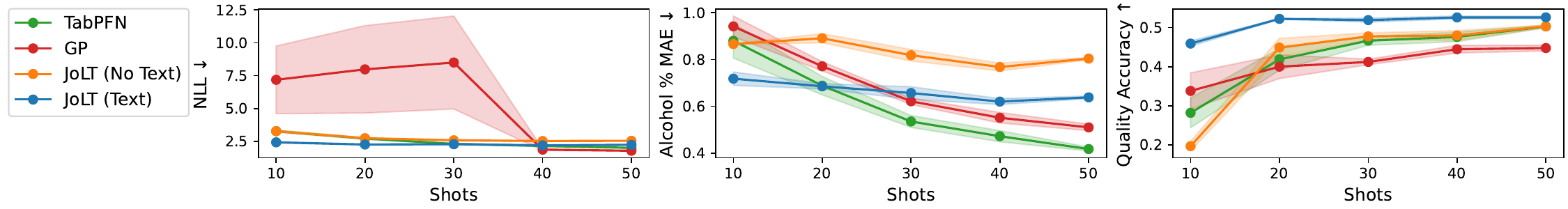}}
\vspace{-4mm}
\caption{Results for predicting two target columns from the Wine Quality dataset \citep{wine_quality_186} as a function of shots when evaluating on 1000 test examples. The first target column is numerical (Alcohol $\%$) using the metric Mean Absolute Error (MAE) and the second target column is categorical (Quality on a scale of 1 to 10) using classification accuracy as the metric. The joint NLL is over both targets. The JoLT methods use the Gemma-2-27B LLM. JoLT (Text) utilized both prefix text $\langle prefix \rangle$ and text from the column headers $X_j, Y_j$, whereas JoLT (No Text) did not. The solid line and dots indicate the mean over 5 seeds which affect the training shot and test example selection and the shaded region shows a confidence interval of one $\sigma$. Tabular results are in \cref{tab:multi_column_results}.}
\label{fig:wine_multi_target}
\end{center}
\vskip -0.35in
\end{figure*}

\textbf{Wine Quality}
The first dataset is the Wine Quality dataset \citep{wine_quality_186} where there are two targets - one numerical (Alcohol $\%$) and one categorical (Wine Quality on the scale of 0 to 10).
This dataset is primarily comprised of numerical features, with the color feature column being the only categorical variable.
A sample JoLT prompt is shown in \cref{app:wine_quality_prompt}.
We compare JoLT to two competitive methods that can make probabilistic predictions - TabPFN \citep{hollmann2025tabpfn, hollmann2025github} and a Gaussian Process (GP).
To produce multiple targets with TabPFN and the GP, we query the models autoregressively with multiple passes.  
We use two variants of the LLM. One that uses both prefix text $\langle prefix \rangle$ and text from the column headers $X_j, Y_j$,
and a no text version that does not.

\cref{fig:wine_multi_target} indicates that JoLT without using textual information performs poorly across the board.
When using text, JoLT has the best classification accuracy, and the best NLL up to 40 shots, but loses out to TabPFN on MAE after 20 shots and to the GP at 30 shots and beyond.
In all cases, TabPFN and the GP improve rapidly with increasing number of shots.
When leveraging text, JoLT outperforms other models in the low-shot setting, but as the amount of training data increases, TabPFN dominates in terms of performance.
Given that this dataset is primarily numerical, it is unsurprising that the advantage of using an LLM-based method reduces as the number of training examples increases.

% Table generated by Excel2LaTeX from sheet 'movies'
\begin{table*}[h]
  \centering
  \caption{\textbf{Movie Results}. JoLT used the Gemma-2-27B LLM which released before the release date of the test movies. Bold indicates the highest results.}
  \label{tab:movies}%
  \vskip 0.05in
  \begin{small}
  \begin{sc}
  \begin{adjustbox}{max width=\textwidth}
    \begin{tabular}{lccccccccccc}
    \toprule
          & \multicolumn{1}{p{4.045em}}{ \centering Rating} & \multicolumn{1}{p{5.135em}}{\centering Adventure} & \multicolumn{1}{p{4.045em}}{ \centering Comedy} & \multicolumn{1}{p{4.045em}}{\centering Family} & \multicolumn{1}{p{4.045em}}{\centering Action} & \multicolumn{1}{p{4.045em}}{\centering Fantasy} & \multicolumn{1}{p{4.045em}}{ Thriller} & \multicolumn{1}{p{4.045em}}{\centering Drama} & \multicolumn{1}{p{4.045em}}{\centering Horror} & \multicolumn{1}{p{4.045em}}{\centering Mean} & \multicolumn{1}{p{4.045em}}{\centering NLL} \\
    Method & \centering (MAE ↓) & \centering (AUC ↑) & \centering (AUC ↑) & \centering (AUC ↑) & \centering (AUC ↑) & \centering (AUC ↑) & \centering (AUC ↑) & (AUC ↑) & (AUC ↑) & (AUC ↑) & ↓ \\
    \midrule
    TabPFN & \textbf{1.09} & 0.55  & 0.59  & 0.55  & 0.64  & 0.57  & 0.32  & 0.55  & 0.52  & 0.54  & 5.59 \\
    JoLT  & \textbf{1.09} & \textbf{0.87} & \textbf{0.94} & \textbf{0.97} & \textbf{0.93} & \textbf{0.94} & \textbf{0.82} & \textbf{0.85} & \textbf{0.99} & \textbf{0.91} & \textbf{4.02} \\
    \bottomrule
    \end{tabular}%
    \end{adjustbox}
    \end{sc}
    \end{small}
\vspace{-4mm}
\end{table*}%

\textbf{Movies}
The second dataset is a subset of the Movies Box Office Dataset (2000-2024) \citep{jilla2024movies} where we predict the movie rating as a continuous value on a scale of 0 to 10, as well as eight binary categorical variables that indicate the movie genre, based on features that include the movie title and worldwide box office revenue.
We only used the 2024 data and split the dataset (89 train, 99 test examples) so that the movies in the test set have a release date after the release date of the Gemma-2 LLM. This was done to ensure that the LLM we use was not pretrained on the test data.
A sample JoLT prompt is shown in \cref{app:movies_prompt}.
The results are shown in \cref{tab:movies}.
When predicting the numerical rating, both JoLT and TabPFN have the same error.
However, JoLT outperforms TabPFN by a large margin in terms of AUC and NLL in predicting the genre attributes of the movies.
This is due to JoLT being able to use the text of the movie title to help predict the binary genre targets, whereas TabPFN is not able to use text columns and hence does only a little better than guessing these targets.

% Table generated by Excel2LaTeX from sheet 'Medals'
\begin{table}[t]
\vspace{-2mm}
  \centering
  \caption{\textbf{Medals Results}. JoLT used the Gemma-2-9B LLM which was released before the Paris Olympics. Values are the mean of 3 seeds. Bold indicates the highest results.}
  \label{tab:multi_column_medals}%
  \begin{small}
  \begin{sc}
  \begin{adjustbox}{max width=\columnwidth}
    \begin{tabular}{lccc}
    \toprule
    Method & \multicolumn{1}{p{9.3em}}{\centering Silver (MAE ↓)} & \multicolumn{1}{p{6.6em}}{\centering Gold (MAE ↓)} & \multicolumn{1}{p{4.2em}}{\centering NLL ↓} \\
    \midrule
    TabPFN & 4.16  & 4.91  & 5.89 \\
    JoLT  & \textbf{3.50} & \textbf{4.80} & \textbf{1.87} \\
    \bottomrule
    \end{tabular}%
    \end{adjustbox}
    \end{sc}
    \end{small}
\vspace{-5mm}
\end{table}%

\textbf{Medals}
The third dataset is the Olympic Games dataset collection \citep{ismail2024olympic} where we use the bronze, silver, and gold medal counts of 10 countries (USA, China, Japan, Australia, France, Netherlands, Great Britain, Italy, Germany, and Canada) from the 1996 to 2020 summer Olympics to train on.
The goal is to predict the silver and gold medal counts for the ten counties at the 2024 Olympics which were held after the release date of the Gemma-2 LLM that JoLT uses.
We treat medal counts as continuous variables for both prediction and NLL computation. 
The bin size for JoLT was selected to be comparable to the bin sizes predicted by TabPFN, ensuring a fair comparison between methods. 
A sample prompt is in \cref{app:medals_prompt}.
Results are shown in \cref{tab:multi_column_medals}.
JoLT has lower MAE and NLL when predicting the 2024 silver and gold medal counts for the 2024 Paris Olympics as it can use the name of the country as context, whereas TabPFN only gets a numerical country label.

\textbf{Summary}: JoLT performs best in low-shot settings due to the ability to utilize user-provided side information and latent knowledge in the LLM.
Performance is superior when dataset features contain rich text content that purely numerical approaches cannot exploit.
\subsection{Side Information Influence}
\begin{figure*}[t]
% \vskip 0.2in
\begin{center}
\centerline{\includegraphics[width=1.0\textwidth]{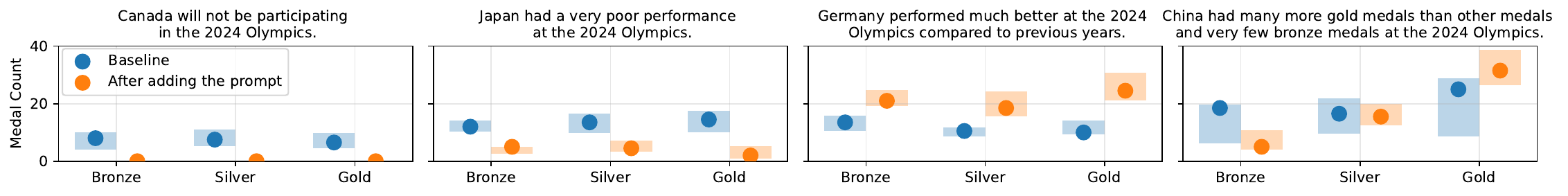}}
\vspace{-4mm}
\caption{Effect of side information on JoLT's predictive distribution for the Medals Dataset. Each subplot corresponds to a different $\langle prefix \rangle$, shown in the title, and represents the distributions for the country mentioned in the $\langle prefix \rangle$. The baseline predictive distribution is in blue, while the distribution after adding side information is in orange. Shaded regions indicate the 25th and 75th percentiles, and markers represent the median values. }

\label{fig:prompt_influence}
\end{center}
\vskip -0.3in
\end{figure*}
In this section, we evaluate the effect of incorporating side textual information on JoLT's predictive distribution.
Using the Medals dataset from \cref{sec:multi_target_results}, we predict the counts for bronze, silver, and gold medals. 
In \cref{fig:prompt_influence}, we apply four different  $\langle prefix \rangle$ values that characterize a country's performance at the 2024 Olympics.
In all cases, the resulting distribution for the selected country shifts to better align with the provided textual information, demonstrating the effectiveness of textual side information in refining JoLT’s predictions.
\subsection{Handling Missing Data}
\label{sec:missing_data_results}
\begin{figure*}[t]
%\vskip 0.2in
\begin{center}
\centerline{\includegraphics[width=1.0\textwidth]{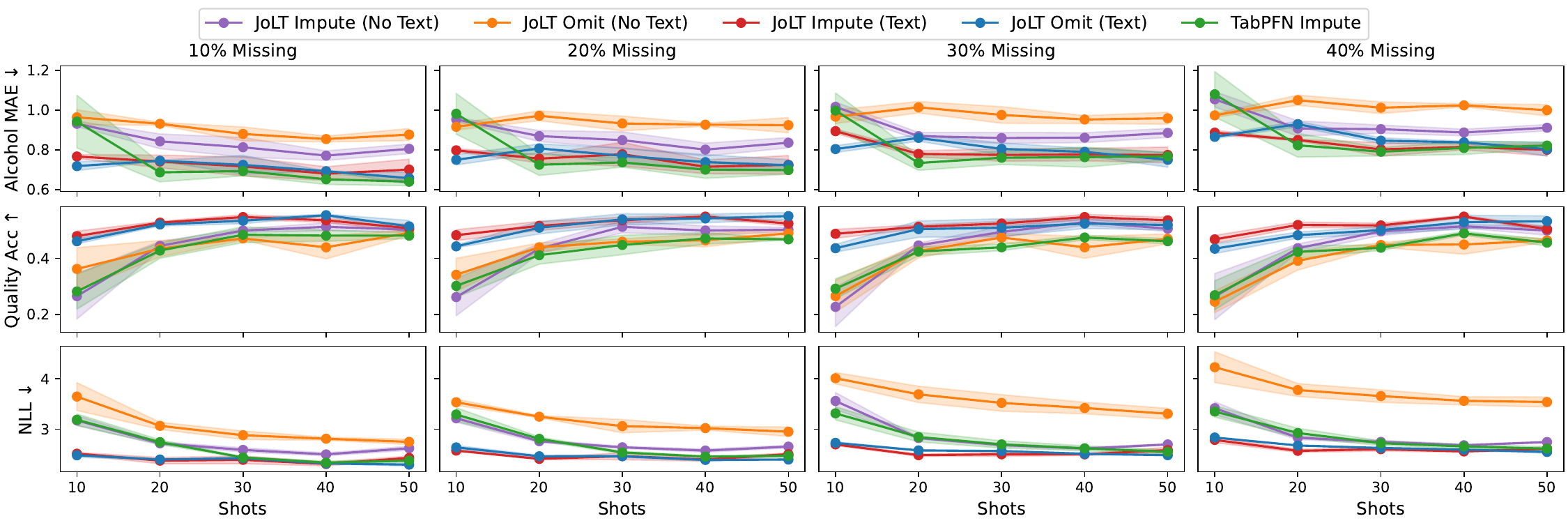}}
\vspace{-4mm}
\caption{Performance metrics for JoLT that uses the gemma-2-27B LLM and TabPFN as a function of shots and percentage of data missing completely-at-random (MCAR) on the multitarget Wine Quality dataset \citep{wine_quality_186}. 200 test examples were used. The solid line and dots indicate the mean over 3 seeds which affect the training shot and test example selection as well as the missing pattern and the shaded region shows a confidence interval of one $\sigma$. The first target column is numerical (Alcohol $\%$) using the metric Mean Absolute Error (MAE) and the second target column is categorical (Wine Quality on a scale of 1 to 10) using classification accuracy as the metric (ACC). The joint negative log-likelihood (NLL) is over both targets. Tabular version is in \cref{tab:wine_missing}.}
\label{fig:wine_missing}
\end{center}
\vskip -0.35in
\end{figure*}
\begin{figure*}[h]
% \vskip 0.2in
\begin{center}
\centerline{\includegraphics[width=1.0\textwidth]{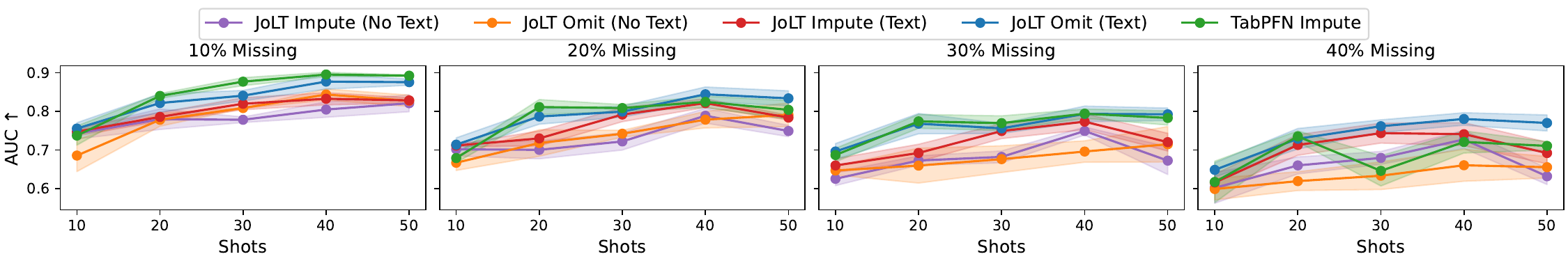}}
\vspace{-4mm}
\caption{AUC for JoLT using the gemma-2-27B LLM and TabPFN as a function of shots and percentage of data missing completely at random (MCAR) on the Car classification dataset \citep{car_evaluation_19}. 200 test examples were used.  The solid line and dots indicate the mean over 3 seeds which affect the training shot and test example selection as well as the missing pattern and the shaded region shows a confidence interval of one $\sigma$. Tabular version is in \cref{tab:car_missing}}
\label{fig:cars_missing}
\end{center}
\vskip -0.3in
\end{figure*}
In these experiments, we compare the \textit{omit} approach to handling missing data as described in \cref{sec:missing_data} to imputing missing data as a preprocessing step.
To impute missing data, we replace missing numerical and categorical values with the column-wise mean and the column-wise mode from the training data, respectfully.
We use two different datasets -- the Wine Quality dataset as used in \cref{sec:multi_target_results} whose features are primarily numerical, and the Car dataset used in \cref{sec:classification_results} whose features are all categorical.

\cref{fig:wine_missing,fig:cars_missing} show the results for the two datasets where the amount of data missing completely-at-random (MCAR) varies from 10$\%$ to 40$\%$ in both the training and test features.
We evaluate four JoLT variants -- omit and impute with and without text, as well TabPFN.
As previously, the variants that do not use side text information perform the worst on both datasets.
Remarkably, for the JoLT variants that use side text information, the omit approach performs almost identically to imputing data on the Wine Quality dataset and exceeds it on the Car dataset.
Interestingly, the opposite is true when no side text is used --- impute performs better than omit.
This supports our argument that the heading text labels on each feature cell helps the LLM to do a better job of knowing what features are missing (or present) and compensate appropriately.
Compared to TabPFN, JoLT omit outperforms at low shot and performs similarly at higher shot.
The exception is at 40$\%$ missing using the Car dataset where JoLT omit outperforms TabPFN by a large margin.

\textbf{Summary}: JoLT implicitly handles missing data, obviating the need to impute as a preprocessing step, which greatly simplifies the data science workflow.
\subsection{Missing Data Imputation}
\label{sec:imputation_results}
\begin{figure}[h]
% \vskip 0.2in
\begin{center}
\centerline{\includegraphics[width=1.0\columnwidth]{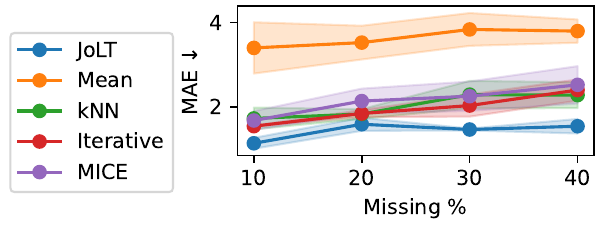}}
\vspace{-5mm}
\caption{\textbf{JoLT imputation.} MAE as a function of $\%$ of missing data (MCAR) on the Paris 2024 Olympic Medals dataset for JoLT that uses the gemma-2-27B and four competive methods. The dots indicate the mean over 3 seeds which affect the missing pattern and the shaded shaded region is a confidence interval of one $\sigma$. Tabular results are in \cref{tab:imputation}.}
\label{fig:imputation}
\end{center}
\vspace{-10mm}
\end{figure}

While JoLT gracefully handles missing data while predicting, it is often required to recover missing feature values.
In this experiment, we demonstrate JoLT's ability to use contextual information and predict multiple values seamlessly to improve imputation performance.
We use the Paris 2024 data from the Olympic Games dataset collection \citep{ismail2024olympic} which lists medals won by each of the 91 nations participating in the event.
We choose this dataset as the event was held after the training cutoff data of the Gemma-2. Hence, it is impossible for the model to have seen this data during pretraining.
In particular, we use the following 4 columns: Country, Gold, Silver, and Bronze.
We then eliminate a fixed percentage of the data completely at random.
The country column cannot be used by existing imputation methods that only deal with numerical data, but can be leveraged by JoLT.
JoLT imputation implementation follows the algorithm described in \cref{sec:imputation}.
An example prompt is shown in \cref{app:imputation_prompt}.
The results are shown in \cref{fig:imputation} where JoLT has significantly lower error than conventional imputation techniques that include MICE~\citep{little2002statistical}, k-Nearest Neighbors~\citep{Troyanskaya2001MissingVE}, mean, and iterative application of a Bayesian Ridge regressor.

\textbf{Summary}: JoLT can outperform standard imputation techniques by leveraging text-based side information about the setting (i.e. the country name, the numerical columns contained medal counts for the 2024 Olympic games, etc.) that is not possible with numerical only methods.
\section{Related Work}
In this section, we survey related methods for tabular prediction and imputation.
For a more complete treatment, refer to \citet{borisov2022deep} for a survey of deep learning methods for tabular data and \citet{fang2024large,lu2024large} for applying LLMs to tabular data.

\textbf{Classification and Multi-target prediction}
TabLLM \citep{hegselmann2023tabllm} fine-tunes an LLM for single target classification only.
It can incorporate side information, but performance is limited in the few-shot setting.
TabPFN \citep{hollmann2025tabpfn} is an effective ICL method for both classification and regression and offers uncertainty in the form of quantiles.
However, it is restricted to handling numerical and categorical data, is unable to incorporate text or side information, cannot predict multiple targets at once, and does not perform well in the low-shot setting.
LLM Processes~\citep{requeima2024llm} support multi-target regression with uncertainty estimates and side information but do not handle classification or heterogeneous targets.
JoLT extends LLM Processes by enabling classification and supporting heterogeneous multiple targets in the tabular data setting.
Carte \citep{kim2024carte} uses a graph-attentional network pretrained across multiple tables, and subsequently fine-tuned on a downstream dataset.
It can incorporate side information while performing single-target regression or classification, but does not provide uncertainty.
Finally, GPs can make multiple target probabilistic predictions on heterogeneous data \citep{moreno2018heterogeneous}, but require training and cannot easily incorporate side information.

\textbf{Handling Missing Data and Data Imputation}
There exists a myriad of widely used imputation methods, including mean, median, and mode imputation, k-Nearest Neighbors~\citep{Troyanskaya2001MissingVE}, MICE~\citep{little2002statistical}, and tree-based methods like MissForest~\citep{Stekhoven2011MissForestN}.
However, they are restricted to using numerical or categorical data, and hence, are unable to exploit side information. Furthermore, they do not provide any estimates of uncertainty or distributions for the imputed values. 
In contrast, Bayesian methods, such as Gaussian Processes (GPs)~\citep{Rasmussen2006Gaussian} and Gaussian Copula~\citep{NelsenRogerB1999Aitc}, offer the advantage of providing uncertainty estimates. 
While Bayesian methods have been extended to handle missing data~\citep{zhaocopula2022,bahram2023gp}, their widespread adoption is hindered by scalability limitations and the challenges associated with incorporating side information effectively.
Recent work has also explored the use of LLMs for data imputation.
\citet{anonymous2024contextdriven} proposes a nearest-neighbor-based imputation method that operates in the embedding space of an LLM.
However, this method does not provide uncertainty estimates for the imputed values. 
\citet{Ding2024DataIU} and \citet{hayat2024claimdataenhancingimputation} fine-tune LLMs to handle missing data and report notable improvements on several downstream tasks. 
To the best of our knowledge, there are no methods that utilize LLMs to perform data imputation in low-shot scenarios.
In contrast, JoLT handles missing data automatically, and if imputed values are required, JoLT can impute missing data with uncertainty, has the ability to incorporate side information, and operate well in the low-shot setting.
\section{Limitations}
Along with the flexibility of LLMs, JoLT inherits their drawbacks.
Maximum context sizes limit the size of tasks we can apply this method to and the amount of textual information we can condition on.
JoLT is significantly more computationally expensive compared to competitive tabular prediction methods.
In particular, both the computational complexity and the maximum context size of the LLM limit the number of training examples and the number of columns in the tabular dataset that can be reasonably processed.
All of the experiments were performed on readily available small to medium sized open source LLMs that have fewer parameters and are generally less capable compared to large open source and proprietary LLMs that are accessed through services.
We expect our results to improve and scale to larger tabular datasets with the use of proprietary LLMs.
%
% Furthermore, we can combine ICL and fine-tuning to generalize to even larger datasets in the future.

\section{Discussion}

In this paper, we introduce JoLT, a novel method for probabilistic predictions on tabular data that leverages LLMs to define joint distributions over heterogeneous data types. 
JoLT distinguishes itself from competitive methods by effectively utilizing side information, providing joint probabilities, and automatically handling missing data,  all without requiring additional training or data preprocessing.
Through extensive experiments, we demonstrate that JoLT excels in low-shot scenarios, particularly when rich text-based side information is available, showcasing its versatility and practicality in real-world applications.
Recently, \citet{agarwal2024many} demonstrated that in-context learning datasets with hundreds or thousands of shots yield performance benefits, indicating that scaling the number of shots might be an interesting direction for the future.
\section*{Impact Statement}
Our work has demonstrated a new and useful ICL approach for generating probabilistic predictions on tabular data.
It has the potential to allow practitioners from fields such as medical and ecological research to more easily employ probabilistic modeling and machine learning.
Like all machine learning technology, there is potential for abuse, and possible consequences from incorrect predictions made with JoLT.
Due to the black-box nature of the method, we do not know the biases in the underlying LLMs used and what effect they may have on JoLT output.
However, LLM researchers are striving to make LLMs more fair and equitable.
An open area of research is whether LLM biases propagate to JoLT predictions and whether de-biasing LLMs helps to fix such an issue.

\section*{Acknowledgements}
James Requeima and David Duvenaud acknowledge funding from the Data Sciences Institute at the University of Toronto and the Vector Institute.
Aliaksandra Shysheya, John Bronskill, and Richard E. Turner are supported by an EPSRC Prosperity Partnership EP/T005386/1 between the EPSRC, Microsoft Research and the University of Cambridge.
Richard E. Turner is also supported by Google, Amazon, ARM, Improbable, and the EPSRC Probabilistic AI Hub (ProbAI, EP/Y028783/1).
\bibliography{main}
\bibliographystyle{icml2025}

%%%%%%%%%%%%%%%%%%%%%%%%%%%%%%%%%%%%%%%%%%%%%%%%%%%%%%%%%%%%%%%%%%%%%%%%%%%%%%%
%%%%%%%%%%%%%%%%%%%%%%%%%%%%%%%%%%%%%%%%%%%%%%%%%%%%%%%%%%%%%%%%%%%%%%%%%%%%%%%
% APPENDIX
%%%%%%%%%%%%%%%%%%%%%%%%%%%%%%%%%%%%%%%%%%%%%%%%%%%%%%%%%%%%%%%%%%%%%%%%%%%%%%%
%%%%%%%%%%%%%%%%%%%%%%%%%%%%%%%%%%%%%%%%%%%%%%%%%%%%%%%%%%%%%%%%%%%%%%%%%%%%%%%
\newpage
\appendix
\onecolumn
\setcounter{figure}{0}
\setcounter{table}{0}
\setcounter{equation}{0}
\setcounter{algorithm}{0}
\renewcommand\thefigure{\thesection.\arabic{figure}} 
\renewcommand\thetable{\thesection.\arabic{table}}
\renewcommand\theequation{\thesection.\arabic{equation}}
\renewcommand\thealgorithm{\thesection.\arabic{algorithm}}

\section{Appendix}
\subsection{Algorithms}
\begin{algorithm*}
\caption{Computing the log probability distribution function of a categorical target $y$}
\label{alg:categorical_logprobs}
\begin{algorithmic}[1]
\Require $\mathcal{M}$: LLM model with vocabulary $\mathcal{V}$; $\mathcal{T}$: tokenizer
\Require $\mathcal{S}$: text string preceding $y$ to condition on; $\mathcal{Y} = \{y_1, \dots y_L\}$ set of possible classes
\State $\mathcal{S}^T = \mathcal{T}(\mathcal{S})$
\Comment{Tokenize $\mathcal{S}$}
\For{$i \gets 1$ to $L$} 
\State $y_i^T \gets \mathcal{T}(y_i)$ 
\Comment{Tokenize $y_i$}
\State $l_i \gets |y_i^T|$ 
\Comment Number of tokens in $y_i^T$
\State logits = $\mathcal{M}(\mathcal{S}^T + y_i^T)$ 
\Comment{Run the LLM model forward, the size of logits is $|\mathcal{S}^T + y_i^T| \times |\mathcal{V}|$}
\For{$j \gets 1$ to $l_i$}
\State y\_logits[j] $\gets$ logits[-$l_i$-2+j, $y_i^T$[j]]
\Comment Logits for $y_i^T$
\EndFor
\State class\_logits[i] $\gets$ y\_logits.sum
\Comment Logits corresponding to class $y_i$
\EndFor
\State y\_log\_pdf $\gets$ \texttt{CrossEntropy}(logits=class\_logits, target=$y$)
\end{algorithmic}
\end{algorithm*}
\begin{algorithm*}
\caption{Computing the log probability distribution function of a numerical target $y$ \citep{requeima2024llm}}
\label{alg:numerical_logprobs}
\begin{algorithmic}[1]
\Require $\mathcal{M}$: LLM model; $\mathcal{T}$: tokenizer
\Require $n$: number of digits after decimal point for $y$
\Require $\mathcal{A} = \{``0", ``1", \ldots, ``9", ``-", ``.", ``\langle t \rangle",  ``\langle s \rangle" \}$: set of allowed characters for $y$
\Require $\mathcal{S}$: text string preceding $y$ to condition on
\State non\_numeric\_mask $\gets$ all\_tokens $\notin \mathcal{T}(\mathcal{A})$ \Comment{Mask of non-numeric tokens}
\State $y^T \gets \mathcal{T}(\texttt{str}(y))$ \Comment{Tokenize $y$}
\State $l \gets |y^T|$ 
\Comment Number of tokens in $y^T$
\State logits = $\mathcal{M}(\mathcal{T}(\mathcal{S}) + y^T)$ \Comment{Run the LLM model forward}
\State y\_logits $\gets$ logits[-$l$-1:-1] \Comment{Logits corresponding to $y$}
\State y\_logits[non\_numeric\_mask] $\gets$ -100 \Comment{Mask out non-numeric tokens}
\State y\_log\_pmf $\gets$ \texttt{CrossEntropy}(logits=y\_logits, targets=$y^T$).sum   \Comment{Probability mass of bin that includes $y$}
\State y\_log\_pdf $\gets$ y\_logp\_mf + $n\log{10}$   \Comment{Convert probability mass to continuous likelihood}
\end{algorithmic}
\end{algorithm*}
\subsection{Sample Prompts}
\label{app:sample_prompts}
In this section, we provide sample prompts for various experiments.
\subsubsection{Wine Quality Experiment Sample Prompt}
\label{app:wine_quality_prompt}
A sample prompt with one training example, plus test example features, and $\langle d \rangle$ = ``:", $\langle s \rangle$ = ``;", and $\langle t \rangle$ = ``\textbackslash n" is:
\begin{tcolorbox}[colback=green!2,colframe=green!40!black]
``The data contains features that determine the quality of wine. Predict the alcohol content and the quality score of each wine based on the features.\textbackslash nfixed\_acidity: 6.2; volatile\_acidity: 0.23; citric\_acid: 0.35; residual\_sugar: 0.7; chlorides: 0.051; free\_sulfur\_dioxide: 24.0; total\_sulfur\_dioxide: 111.0; density: 0.992; pH: 3.37; sulphates: 0.43; 
color: white; alcohol: 11.0; quality: 3\textbackslash nfixed\_acidity: 9.9; volatile\_acidity: 0.49; citric\_acid: 0.23; residual\_sugar: 2.4; chlorides: 0.087; free\_sulfur\_dioxide: 19.0; total\_sulfur\_dioxide: 115.0; density: 0.995; pH: 2.77; sulphates: 0.44; color: white;"
\end{tcolorbox}
\subsubsection{Movies Box Office Sample Prompt}
\label{app:movies_prompt}
A sample prompt with one training example, plus test example features, and $\langle d \rangle$ = ``:", $\langle s \rangle$ = ``;", and $\langle t \rangle$ = ``\textbackslash n" is:
\begin{tcolorbox}[colback=green!2,colframe=green!40!black]
``Each example contains 10 columns: Movie Name, Revenue in Millions of Dollars, Rating, and 8 genre tags (Adventure, Comedy, Family, Action, Fantasy, Thriller, Drama, and Horror). Predict the movie rating and genre tags.\textbackslash nMovie Name:Kung Fu Panda 4;ID:36;Revenue in \$Millions:547.7;Rating:7.1;Adventure:No;Comedy:No;Family:Yes;Action:Yes;Fantasy:Yes;Thriller:No;Drama:No;Horror:\\No\textbackslash nMovie Name:Speak No Evil;ID:120;Revenue in \$Millions:76.8;"
\end{tcolorbox}
\subsubsection{Olympic Games Dataset Prompt}
\label{app:medals_prompt}
A sample prompt with one training example, plus test example features, and $\langle d \rangle$ = ``:", $\langle s \rangle$ = ``;", and $\langle t \rangle$ = ``\textbackslash n" is:
\begin{tcolorbox}[colback=green!2,colframe=green!40!black]
``Each example contains five columns: Olympic Year, Country, Bronze Medal Count, Silver Medal Count, and Gold Medal Count that describe what type and how many medals a country won at the Olympic games that year. Predict the number of silver and gold medals won by that country in that year.\textbackslash nOlympic Year:2020;Country:Netherlands;Bronze Medal Count:14;Silver Medal Count:12;Gold Medal Count:10\textbackslash nOlympic Year:2024;Country:USA;Bronze Medal Count:42;'"
\end{tcolorbox}
\subsubsection{Imputation Prompt}
\label{app:imputation_prompt}
A sample prompt with one training example, plus test example features, and $\langle d \rangle$ = ``:", $\langle s \rangle$ = ``;", and $\langle t \rangle$ = ``\textbackslash n" is:
\begin{tcolorbox}[colback=green!2,colframe=green!40!black]
``Each example contains four columns: Country, Silver Medal Count, Bronze Medal Count, and Gold Medal Count that describe what type and how many medals a country won at the Paris 2024 olympics.\textbackslash nCountry:Thailand;Silver Medal Count:3;Bronze Medal Count:2;Gold Medal Count:1\textbackslash nCountry:Slovakia;Silver Medal Count:0;Bronze Medal Count:1;"
\end{tcolorbox}
\subsection{Datasets}
\label{app:datasets}
\paragraph{Classification}

For the few-shot classification experiments, we utilize the nine datasets introduced in \citet{hegselmann2023tabllm}. The serialized versions of these datasets are obtained from \citet{hegselmann2023github}. These datasets are:
\begin{itemize}
    \item \textbf{Bank}~\citep{bank_marketing_222} contains records relevant to a direct marketing campaigns (phone calls) of a Portuguese banking institution. The classification goal is to predict if the client will subscribe to a term deposit (binary classification). It consists of $45,211$ rows and $16$ features; $5,289$ labels are positive. 
    
    \item \textbf{Blood}~\citep{blood_transfusion_service_center_176} contains $748$ donor records from the Blood Transfusion Service Center in Taiwan. The classification task is to predict whether a donor returned for another blood donation.
    
    \item \textbf{Calhousing}~\citep{pace1997sparse} has entries of houses found in a given California district and some summary stats about them based on the 1990 U.S. census data. Following \citep{hegselmann2023github}, the goal is to predict whether the house value is below or above the median in its district.
 
    \item \textbf{Cars}~\citep{car_evaluation_19} contains records of various cars, characterized by six attributes. The task is a multiclass classification problem to evaluate the state of each car.
    
    \item \textbf{Creditg}~\citep{creditg_1994} contains $1000$ records, each described by $20$ attributes, and the task is to classify individuals as either good or bad credit risks. Of these records, $700$ are classified as good credit risks.
    
    \item \textbf{Diabetes}~\citep{smith1988diabetes}: originally from the National Institute of Diabetes and Digestive and Kidney Diseases, this dataset aims to diagnostically predict whether a patient has diabetes based on specific diagnostic measurements. Among the records, $268$ cases are positive for diabetes.
    
    \item \textbf{Heart}~\citep{detrano1989international} combines records from four hospitals in Cleveland, Hungary, Switzerland, and Long Beach V. The task is to predict the presence of heart disease in patients. Among the $918$ patients, $508$ are diagnosed as positive for heart disease.
    
    \item \textbf{Income}~\citep{income_1996} contains records of $48,842$ individuals with $12$ attributes collected in the 1994 U.S. Census. The classification task is to predict 
    whether annual income of an individual exceeds \$50K. The dataset has $11,687$ positive labels.
    
    \item \textbf{Jungle}~\citep{jungle_2014} consists of $44,819$ endgame positions from Jungle Chess. Each position is described by $6$ attributes, and the task is to predict whether the white player wins. Among the records, $23,062$ are positive outcomes for the white player.
\end{itemize}

\paragraph{Multi-target prediction}

For the multi-target prediction in \cref{sec:multi_target_results}, we use the following datasets: 

\begin{itemize}
    \item \textbf{Wine Quality}~\citep{wine_quality_186} contains records of Portuguese wines, each characterized by 11 physiochemical attributes. The original task is to predict wine quality on a scale from 0 to 10. To evaluate multi-target prediction capabilities, we modified the task to predict both the wine quality (categorical) and alcohol percentage (numerical).

    \item \textbf{Movies Box office Dataset (2000-2024)}~\citep{jilla2024movies} provides a comprehensive analysis of global box office performance from 2000 to 2024. Each row represents a movie and includes attributes such as release year, genres, production budget, worldwide gross, and additional descriptive features. The dataset contains $4955$ movies. 

    \item \textbf{Olympic Games (1994-2024)}~\citep{ismail2024olympic} is a database of medals awarded at the Summer and Winter Olympic Games from 1994 to 2024. Each table in the database corresponds to a specific Olympic Games, detailing medal counts by participating countries. Each row includes the country code and the number of gold, silver, and bronze medals won by the respective country.

\end{itemize}
In \cref{sec:missing_data_results}, we evaluate using the \textbf{Cars} and \textbf{Wine Quality} datasets, while in \cref{sec:imputation_results}, we use a subset of the \textbf{Olympic Games} dataset.

\subsection{Uncertainty Example}
\begin{figure}[h]
\vskip 0.2in
\begin{center}
\centerline{\includegraphics[width=0.8\columnwidth]{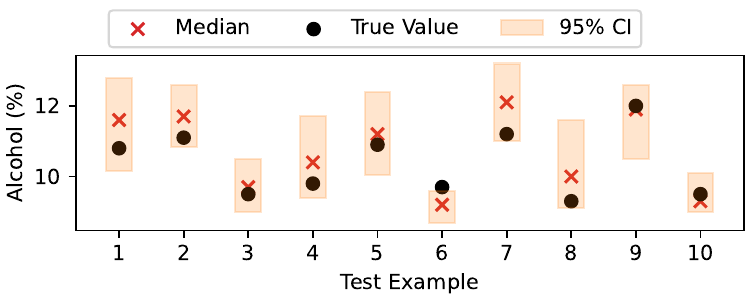}}
\vspace{-2mm}
\caption{\textbf{JoLT predictions with uncertainty.} Median point estimates and 95$\%$ confidence intervals for predictions made by an JoLT using 100 samples with top-$p=$0.9 on the first 10 test examples on the wine quality dataset  \citep{wine_quality_186}.}
\label{fig:uncertainty}
\end{center}
\vskip -0.2in
\end{figure}

\subsection{Additional Experimental Results}
In this section, we provide tabular versions of the results that were presented as plots in the main paper.
    For any experiment with a regression target, we use \textit{top}-1 sampling. The one exception is in the Movies experiment, where we use 50 samples and take the median as the point estimate.
% Table generated by Excel2LaTeX from sheet 'Paper Version'
\begin{table*}[htbp]
  \centering
  \caption{\textbf{Classification}. AUC as a function of shot for three JoLT configurations and three competitive methods. Values are the mean and 95$\%$ confidence interval (CI) over 5 seeds that affect the training shot selection. Due to limited computational resources values at 16 and 32 shots with 0 CI use only a single seed. Competitive data from \citep{hegselmann2023tabllm}.}
  \label{tab:classification_results}%
  \vskip 0.05in
  \begin{small}
  \begin{sc}
  \begin{adjustbox}{max width=\textwidth}
    \begin{tabular}{rcccccc}
    \toprule
          &       & \multicolumn{5}{c}{\textbf{Shot}} \\
\cmidrule{3-7}    \multicolumn{1}{l}{\textbf{Dataset}} & \textbf{Method} & \textbf{0} & \textbf{4} & \textbf{8} & \textbf{16} & \textbf{32} \\
    \midrule
          & XGBoost & -     & 0.5$\pm$0.00 & 0.56$\pm$0.08 & 0.68$\pm$0.04 & 0.76$\pm$0.03 \\
          & TabPFN & -     & 0.59$\pm$0.12 & 0.66$\pm$0.07 & 0.69$\pm$0.02 & 0.76$\pm$0.03 \\
    \multicolumn{1}{l}{bank} & TabLLM & 0.63$\pm$0.01 & 0.59$\pm$0.09 & 0.64$\pm$0.04 & 0.65$\pm$0.04 & 0.64$\pm$0.05 \\
          & JoLT (Gemma-2-2B) & 0.46$\pm$0.00 & 0.62$\pm$0.05 & 0.62$\pm$0.05 & 0.57$\pm$0.05 & 0.52$\pm$0.09 \\
          & JoLT (Gemma-2-27B) & 0.61$\pm$0.00 & 0.72$\pm$0.01 & 0.72$\pm$0.01 & 0.71$\pm$0.01 & 0.73$\pm$0.00 \\
          & JoLT (Qwen-2.5-72B) & 0.73$\pm$0.00 & 0.83$\pm$0.01 & 0.81$\pm$0.02 & 0.77$\pm$0.00 & 0.73$\pm$0.00 \\
    \midrule
          & XGBoost & -     & 0.5$\pm$0.00 & 0.58$\pm$0.06 & 0.66$\pm$0.04 & 0.67$\pm$0.05 \\
          & TabPFN & -     & 0.52$\pm$0.07 & 0.64$\pm$0.04 & 0.67$\pm$0.01 & 0.7$\pm$0.04 \\
    \multicolumn{1}{l}{blood} & TabLLM & 0.61$\pm$0.04 & 0.58$\pm$0.08 & 0.66$\pm$0.03 & 0.66$\pm$0.06 & 0.68$\pm$0.04 \\
          & JoLT (Gemma-2-2B) & 0.56$\pm$0.00 & 0.62$\pm$0.08 & 0.58$\pm$0.04 & 0.64$\pm$0.06 & 0.59$\pm$0.05 \\
          & JoLT (Gemma-2-27B) & 0.70$\pm$0.00 & 0.72$\pm$0.03 & 0.68$\pm$0.03 & 0.73$\pm$0.02 & 0.71$\pm$0.05 \\
          & JoLT (Qwen-2.5-72B) & 0.64$\pm$0.00 & 0.71$\pm$0.04 & 0.71$\pm$0.02 & 0.73$\pm$0.02 & 0.72$\pm$0.03 \\
    \midrule
          & XGBoost & -     & 0.5$\pm$0.00 & 0.62$\pm$0.09 & 0.74$\pm$0.03 & 0.79$\pm$0.04 \\
          & TabPFN & -     & 0.63$\pm$0.11 & 0.63$\pm$0.10 & 0.8$\pm$0.03 & 0.85$\pm$0.03 \\
    \multicolumn{1}{l}{calhousing} & TabLLM & 0.61$\pm$0.01 & 0.63$\pm$0.04 & 0.6$\pm$0.06 & 0.7$\pm$0.07 & 0.77$\pm$0.07 \\
          & JoLT (Gemma-2-2B) & 0.45$\pm$0.00 & 0.72$\pm$0.01 & 0.66$\pm$0.09 & 0.76$\pm$0.04 & 0.78$\pm$0.04 \\
          & JoLT (Gemma-2-27B) & 0.64$\pm$0.00 & 0.83$\pm$0.01 & 0.82$\pm$0.04 & 0.85$\pm$0.02 & 0.85$\pm$0.01 \\
          & JoLT (Qwen-2.5-72B) & 0.64$\pm$0.00 & 0.83$\pm$0.01 & 0.80$\pm$0.04 & 0.84$\pm$0.01 & 0.84$\pm$0.01 \\
    \midrule
          & XGBoost & -     & 0.5$\pm$0.00 & 0.59$\pm$0.04 & 0.7$\pm$0.07 & 0.82$\pm$0.03 \\
          & TabPFN & -     & 0.64$\pm$0.05 & 0.75$\pm$0.04 & 0.87$\pm$0.04 & 0.92$\pm$0.02 \\
    \multicolumn{1}{l}{car} & TabLLM & 0.82$\pm$0.02 & 0.83$\pm$0.03 & 0.85$\pm$0.03 & 0.86$\pm$0.03 & 0.91$\pm$0.02 \\
          & JoLT (Gemma-2-2B) & 0.73$\pm$0.00 & 0.84$\pm$0.02 & 0.79$\pm$0.02 & 0.79$\pm$0.04 & 0.74$\pm$0.04 \\
          & JoLT (Gemma-2-27B) & 0.82$\pm$0.00 & 0.89$\pm$0.01 & 0.90$\pm$0.01 & 0.91$\pm$0.01 & 0.94$\pm$0.01 \\
          & JoLT (Qwen-2.5-72B) & 0.86$\pm$0.00 & 0.89$\pm$0.04 & 0.88$\pm$0.02 & 0.89$\pm$0.04 & 0.94$\pm$0.01 \\
    \midrule
          & XGBoost & -     & 0.5$\pm$0.00 & 0.51$\pm$0.06 & 0.59$\pm$0.04 & 0.66$\pm$0.03 \\
          & TabPFN & -     & 0.58$\pm$0.07 & 0.59$\pm$0.03 & 0.64$\pm$0.05 & 0.69$\pm$0.06 \\
    \multicolumn{1}{l}{creditg} & TabLLM & 0.53$\pm$0.04 & 0.69$\pm$0.04 & 0.66$\pm$0.04 & 0.66$\pm$0.04 & 0.72$\pm$0.05 \\
          & JoLT (Gemma-2-2B) & 0.52$\pm$0.00 & 0.52$\pm$0.03 & 0.53$\pm$0.06 & 0.55$\pm$0.04 & 0.50$\pm$0.04 \\
          & JoLT (Gemma-2-27B) & 0.48$\pm$0.00 & 0.55$\pm$0.02 & 0.54$\pm$0.04 & 0.56$\pm$0.04 & 0.53$\pm$0.01 \\
          & JoLT (Qwen-2.5-72B) & 0.60$\pm$0.00 & 0.56$\pm$0.03 & 0.57$\pm$0.05 & 0.60$\pm$0.08 & 0.65$\pm$0.05 \\
    \midrule
          & XGBoost & -     & 0.5$\pm$0.00 & 0.59$\pm$0.14 & 0.72$\pm$0.06 & 0.69$\pm$0.07 \\
          & TabPFN & -     & 0.61$\pm$0.11 & 0.67$\pm$0.10 & 0.71$\pm$0.06 & 0.77$\pm$0.03 \\
    \multicolumn{1}{l}{diabetes} & TabLLM & 0.68$\pm$0.05 & 0.61$\pm$0.08 & 0.63$\pm$0.07 & 0.69$\pm$0.06 & 0.68$\pm$0.04 \\
          & JoLT (Gemma-2-2B) & 0.62$\pm$0.00 & 0.73$\pm$0.06 & 0.71$\pm$0.06 & 0.76$\pm$0.02 & 0.73$\pm$0.06 \\
          & JoLT (Gemma-2-27B) & 0.82$\pm$0.00 & 0.81$\pm$0.02 & 0.79$\pm$0.01 & 0.80$\pm$0.02 & 0.80$\pm$0.02 \\
          & JoLT (Qwen-2.5-72B) & 0.78$\pm$0.00 & 0.83$\pm$0.02 & 0.82$\pm$0.02 & 0.82$\pm$0.02 & 0.82$\pm$0.02 \\
    \midrule
          & XGBoost & -     & 0.5$\pm$0.00 & 0.55$\pm$0.12 & 0.84$\pm$0.06 & 0.88$\pm$0.04 \\
          & TabPFN & -     & 0.84$\pm$0.05 & 0.88$\pm$0.04 & 0.87$\pm$0.05 & 0.91$\pm$0.02 \\
    \multicolumn{1}{l}{heart} & TabLLM & 0.54$\pm$0.04 & 0.76$\pm$0.12 & 0.83$\pm$0.04 & 0.87$\pm$0.04 & 0.87$\pm$0.05 \\
          & JoLT (Gemma-2-2B) & 0.64$\pm$0.00 & 0.74$\pm$0.01 & 0.80$\pm$0.04 & 0.72$\pm$0.08 & 0.65$\pm$0.09 \\
          & JoLT (Gemma-2-27B) & 0.74$\pm$0.00 & 0.82$\pm$0.02 & 0.85$\pm$0.01 & 0.87$\pm$0.01 & 0.88$\pm$0.01 \\
          & JoLT (Qwen-2.5-72B) & 0.89$\pm$0.00 & 0.87$\pm$0.01 & 0.88$\pm$0.01 & 0.89$\pm$0.01 & 0.90$\pm$0.02 \\
    \midrule
          & XGBoost & -     & 0.5$\pm$0.00 & 0.59$\pm$0.05 & 0.77$\pm$0.02 & 0.79$\pm$0.03 \\
          & TabPFN & -     & 0.73$\pm$0.07 & 0.71$\pm$0.08 & 0.76$\pm$0.08 & 0.8$\pm$0.04 \\
    \multicolumn{1}{l}{income} & TabLLM & 0.84$\pm$0.00 & 0.84$\pm$0.01 & 0.84$\pm$0.02 & 0.84$\pm$0.04 & 0.84$\pm$0.01 \\
          & JoLT (Gemma-2-2B) & 0.82$\pm$0.00 & 0.82$\pm$0.02 & 0.82$\pm$0.01 & 0.83$\pm$0.02 & 0.83$\pm$0.01 \\
          & JoLT (Gemma-2-27B) & 0.86$\pm$0.00 & 0.85$\pm$0.00 & 0.85$\pm$0.01 & 0.85$\pm$0.01 & 0.85$\pm$0.00 \\
          & JoLT (Qwen-2.5-72B) & 0.83$\pm$0.00 & 0.86$\pm$0.00 & 0.85$\pm$0.01 & 0.86$\pm$0.00 & 0.86$\pm$0.00 \\
    \midrule
          & XGBoost & -     & 0.5$\pm$0.00 & 0.58$\pm$0.06 & 0.72$\pm$0.04 & 0.78$\pm$0.03 \\
          & TabPFN & -     & 0.65$\pm$0.07 & 0.72$\pm$0.04 & 0.71$\pm$0.06 & 0.78$\pm$0.02 \\
    \multicolumn{1}{l}{jungle} & TabLLM & 0.6$\pm$0.00 & 0.64$\pm$0.01 & 0.64$\pm$0.02 & 0.65$\pm$0.03 & 0.71$\pm$0.02 \\
          & JoLT (Gemma-2-2B) & 0.67$\pm$0.00 & 0.60$\pm$0.02 & 0.59$\pm$0.04 & 0.55$\pm$0.04 & 0.61$\pm$0.04 \\
          & JoLT (Gemma-2-27B) & 0.62$\pm$0.00 & 0.62$\pm$0.01 & 0.63$\pm$0.01 & 0.63$\pm$0.02 & 0.64$\pm$0.01 \\
          & JoLT (Qwen-2.5-72B) & 0.62$\pm$0.00 & 0.62$\pm$0.01 & 0.62$\pm$0.01 & 0.61$\pm$0.03 & 0.64$\pm$0.01 \\
    \bottomrule
    \end{tabular}%
    \end{adjustbox}
  \end{sc}
  \end{small}
  \vskip -0.1in
\end{table*}%

\begin{table*}[htbp]
  \centering
  \caption{\textbf{Multi-target Prediction}. Results for predicting two target columns from the Wine Quality dataset \citep{wine_quality_186} as a function of shots. 1000 test examples were used. The first target column is numerical (Alcohol $\%$) using the metric Mean Absolute Error (MAE) and the second target column is categorical (Quality on a scale of 1 to 10) using classification accuracy as the metric. The joint NLL is over both targets. The LLMP methods use the Gemma-2-27B LLM. LLMP (Text) utilized both prefix text $\langle prefix \rangle$ and text from the column headers $X_j, Y_j$, whereas LLMP (No Text) did not. Values are the mean and 95$\%$ confidence interval (CI) over 5 seeds that affect the training shot and test example selection.}
  \label{tab:multi_column_results}%
  \vskip 0.1in
  \begin{small}
  \begin{sc}
  \begin{adjustbox}{max width=\textwidth}
    \begin{tabular}{rclllll}
    \toprule
          &       & \multicolumn{5}{c}{\textbf{Shots}} \\
\cmidrule{3-7}    \multicolumn{1}{l}{\textbf{Method}} & \textbf{Metric} & \multicolumn{1}{c}{\textbf{10}} & \multicolumn{1}{c}{\textbf{20}} & \multicolumn{1}{c}{\textbf{30}} & \multicolumn{1}{c}{\textbf{40}} & \multicolumn{1}{c}{\textbf{50}} \\
    \midrule
          & MAE↓  & 0.880$\pm$0.144 & 0.688$\pm$0.077 & 0.535$\pm$0.050 & 0.472$\pm$0.047 & 0.418$\pm$0.017 \\
    \multicolumn{1}{l}{TabPFN} & ACC↑  & 0.282$\pm$0.074 & 0.419$\pm$0.054 & 0.466$\pm$0.023 & 0.476$\pm$0.023 & 0.503$\pm$0.007 \\
          & NLL↓  & 3.281$\pm$0.149 & 2.724$\pm$0.135 & 2.314$\pm$0.060 & 2.159$\pm$0.077 & 2.013$\pm$0.030 \\
    \midrule
          & MAE↓  & 0.942$\pm$0.044 & 0.771$\pm$0.020 & 0.622$\pm$0.014 & 0.551$\pm$0.023 & 0.510$\pm$0.014 \\
    \multicolumn{1}{l}{GP} & ACC↑  & 0.338$\pm$0.046 & 0.400$\pm$0.030 & 0.412$\pm$0.007 & 0.445$\pm$0.010 & 0.448$\pm$0.007 \\
          & NLL↓  & 7.185$\pm$2.573 & 7.980$\pm$3.322 & 8.499$\pm$3.532 & 1.879$\pm$0.068 & 1.795$\pm$0.021 \\
    \midrule
          & MAE↓  & 0.866$\pm$0.005 & 0.891$\pm$0.038 & 0.818$\pm$0.049 & 0.768$\pm$0.033 & 0.804$\pm$0.011 \\
    \multicolumn{1}{l}{LLMP (No text)} & ACC↑  & 0.197$\pm$0.024 & 0.449$\pm$0.046 & 0.477$\pm$0.018 & 0.480$\pm$0.024 & 0.503$\pm$0.006 \\
          & NLL↓  & 3.285$\pm$0.143 & 2.747$\pm$0.086 & 2.589$\pm$0.048 & 2.534$\pm$0.032 & 2.552$\pm$0.029 \\
    \midrule
          & MAE↓  & 0.718$\pm$0.056 & 0.686$\pm$0.022 & 0.657$\pm$0.053 & 0.620$\pm$0.025 & 0.638$\pm$0.011 \\
    \multicolumn{1}{l}{LLMP (Text)} & ACC↑  & 0.459$\pm$0.013 & 0.522$\pm$0.002 & 0.519$\pm$0.010 & 0.526$\pm$0.007 & 0.526$\pm$0.006 \\
          & NLL↓  & 2.434$\pm$0.101 & 2.268$\pm$0.048 & 2.293$\pm$0.084 & 2.205$\pm$0.057 & 2.248$\pm$0.019 \\
    \bottomrule
    \end{tabular}%
    \end{adjustbox}
  \end{sc}
  \end{small}
  \vskip -0.1in
\end{table*}%

% Table generated by Excel2LaTeX from sheet 'missing multi-column paper new'
\begin{table*}[htbp]
  \centering
  \caption{\textbf{Missing Data Handling on Wine Quality}. Performance metrics for JoLT that uses the gemma-2-27B LLM and TabPFN as a function of shots and percentage of data missing completely-at-random (MCAR) on the multitarget Wine Quality dataset \citep{wine_quality_186}. 200 test examples were used. Values are the mean and 95$\%$ confidence interval (CI) over 5 seeds that affect the training and test shot selection and the missing pattern. The first target column is numerical (Alcohol $\%$) using the metric Mean Absolute Error (MAE) and the second target column is categorical (Wine Quality on a scale of 1 to 10) using classification accuracy as the metric (ACC). The joint negative log-likelihood (NLL) is over both targets.}
  \label{tab:wine_missing}%
  \vskip 0.1in
  \begin{tiny}
  \begin{sc}
  \begin{adjustbox}{max width=\textwidth}
    \begin{tabular}{crclllll}
    \toprule
          &       &       & \multicolumn{5}{c}{\textbf{Shots}} \\
\cmidrule{3-8}    \textbf{Missing \%} & \multicolumn{1}{c}{\textbf{Method}} & \textbf{Metric} & \multicolumn{1}{c}{\textbf{10}} & \multicolumn{1}{c}{\textbf{20}} & \multicolumn{1}{c}{\textbf{30}} & \multicolumn{1}{c}{\textbf{40}} & \multicolumn{1}{c}{\textbf{50}} \\
    \midrule
          &       & MAE   & 0.943$\pm$0.053 & 0.875$\pm$0.030 & 0.832$\pm$0.081 & 0.809$\pm$0.026 & 0.824$\pm$0.035 \\
          & \multicolumn{1}{l}{LLMP Impute (No Text)} & ACC   & 0.309$\pm$0.120 & 0.475$\pm$0.048 & 0.480$\pm$0.014 & 0.470$\pm$0.051 & 0.473$\pm$0.023 \\
          &       & NLL   & 3.420$\pm$0.428 & 2.746$\pm$0.025 & 2.595$\pm$0.100 & 2.608$\pm$0.036 & 2.570$\pm$0.078 \\
\cmidrule{2-8}          &       & MAE   & 0.955$\pm$0.051 & 0.931$\pm$0.015 & 0.880$\pm$0.069 & 0.854$\pm$0.029 & 0.877$\pm$0.059 \\
          & \multicolumn{1}{l}{LLMP Omit (No Text)} & ACC   & 0.370$\pm$0.100 & 0.435$\pm$0.058 & 0.472$\pm$0.012 & 0.440$\pm$0.081 & 0.492$\pm$0.019 \\
          &       & NLL   & 3.716$\pm$0.377 & 3.067$\pm$0.139 & 2.882$\pm$0.166 & 2.812$\pm$0.058 & 2.750$\pm$0.100 \\
\cmidrule{2-8}          &       & MAE   & 0.718$\pm$0.038 & 0.769$\pm$0.027 & 0.760$\pm$0.106 & 0.707$\pm$0.022 & 0.690$\pm$0.059 \\
    10    & \multicolumn{1}{l}{LLMP Impute (Text)} & ACC   & 0.458$\pm$0.030 & 0.527$\pm$0.023 & 0.532$\pm$0.013 & 0.532$\pm$0.013 & 0.530$\pm$0.020 \\
          &       & NLL   & 2.458$\pm$0.058 & 2.409$\pm$0.043 & 2.428$\pm$0.093 & 2.347$\pm$0.020 & 2.356$\pm$0.041 \\
\cmidrule{2-8}          &       & MAE   & 0.718$\pm$0.045 & 0.745$\pm$0.005 & 0.724$\pm$0.093 & 0.692$\pm$0.001 & 0.657$\pm$0.057 \\
          & \multicolumn{1}{l}{LLMP Omit (Text)} & ACC   & 0.462$\pm$0.019 & 0.522$\pm$0.011 & 0.535$\pm$0.021 & 0.555$\pm$0.007 & 0.515$\pm$0.042 \\
          &       & NLL   & 2.484$\pm$0.077 & 2.401$\pm$0.018 & 2.424$\pm$0.065 & 2.322$\pm$0.010 & 2.299$\pm$0.053 \\
\cmidrule{2-8}          &       & MAE   & 0.942$\pm$0.260 & 0.686$\pm$0.094 & 0.693$\pm$0.047 & 0.652$\pm$0.052 & 0.639$\pm$0.040 \\
          & \multicolumn{1}{l}{TabPFN} & ACC   & 0.282$\pm$0.123 & 0.428$\pm$0.055 & 0.485$\pm$0.055 & 0.482$\pm$0.039 & 0.482$\pm$0.021 \\
          &       & NLL   & 3.188$\pm$0.227 & 2.745$\pm$0.090 & 2.441$\pm$0.028 & 2.339$\pm$0.060 & 2.379$\pm$0.056 \\
    \midrule
          &       & MAE   & 0.956$\pm$0.081 & 0.945$\pm$0.025 & 0.877$\pm$0.071 & 0.867$\pm$0.040 & 0.858$\pm$0.064 \\
          & \multicolumn{1}{l}{LLMP Impute (No Text)} & ACC   & 0.284$\pm$0.091 & 0.467$\pm$0.049 & 0.487$\pm$0.019 & 0.473$\pm$0.072 & 0.495$\pm$0.012 \\
          &       & NLL   & 3.558$\pm$0.508 & 2.852$\pm$0.052 & 2.643$\pm$0.087 & 2.654$\pm$0.028 & 2.607$\pm$0.073 \\
\cmidrule{2-8}          &       & MAE   & 0.936$\pm$0.041 & 0.972$\pm$0.048 & 0.933$\pm$0.073 & 0.927$\pm$0.014 & 0.924$\pm$0.075 \\
          & \multicolumn{1}{l}{LLMP Omit (No Text)} & ACC   & 0.351$\pm$0.088 & 0.440$\pm$0.030 & 0.460$\pm$0.020 & 0.465$\pm$0.045 & 0.490$\pm$0.030 \\
          &       & NLL   & 3.714$\pm$0.318 & 3.248$\pm$0.044 & 3.062$\pm$0.258 & 3.022$\pm$0.074 & 2.951$\pm$0.181 \\
\cmidrule{2-8}          &       & MAE   & 0.781$\pm$0.066 & 0.828$\pm$0.059 & 0.769$\pm$0.098 & 0.743$\pm$0.046 & 0.766$\pm$0.055 \\
    20    & \multicolumn{1}{l}{LLMP Impute (Text)} & ACC   & 0.455$\pm$0.023 & 0.528$\pm$0.042 & 0.520$\pm$0.035 & 0.528$\pm$0.019 & 0.540$\pm$0.005 \\
          &       & NLL   & 2.574$\pm$0.086 & 2.492$\pm$0.053 & 2.474$\pm$0.107 & 2.409$\pm$0.025 & 2.451$\pm$0.009 \\
\cmidrule{2-8}          &       & MAE   & 0.750$\pm$0.051 & 0.807$\pm$0.044 & 0.770$\pm$0.077 & 0.738$\pm$0.074 & 0.722$\pm$0.056 \\
          & \multicolumn{1}{l}{LLMP Omit (Text)} & ACC   & 0.443$\pm$0.010 & 0.510$\pm$0.044 & 0.540$\pm$0.039 & 0.543$\pm$0.031 & 0.552$\pm$0.025 \\
          &       & NLL   & 2.641$\pm$0.073 & 2.465$\pm$0.042 & 2.465$\pm$0.044 & 2.393$\pm$0.039 & 2.400$\pm$0.018 \\
\cmidrule{2-8}          &       & MAE   & 0.982$\pm$0.201 & 0.725$\pm$0.105 & 0.737$\pm$0.047 & 0.700$\pm$0.085 & 0.698$\pm$0.042 \\
          & \multicolumn{1}{l}{TabPFN} & ACC   & 0.302$\pm$0.080 & 0.412$\pm$0.063 & 0.448$\pm$0.068 & 0.472$\pm$0.040 & 0.468$\pm$0.005 \\
          &       & NLL   & 3.292$\pm$0.264 & 2.810$\pm$0.074 & 2.541$\pm$0.053 & 2.459$\pm$0.026 & 2.475$\pm$0.070 \\
    \midrule
          &       & MAE   & 0.958$\pm$0.074 & 0.956$\pm$0.016 & 0.907$\pm$0.053 & 0.903$\pm$0.030 & 0.884$\pm$0.072 \\
          & \multicolumn{1}{l}{LLMP Impute (No Text)} & ACC   & 0.266$\pm$0.083 & 0.455$\pm$0.045 & 0.497$\pm$0.016 & 0.473$\pm$0.056 & 0.495$\pm$0.024 \\
          &       & NLL   & 3.515$\pm$0.441 & 2.888$\pm$0.067 & 2.684$\pm$0.072 & 2.679$\pm$0.053 & 2.648$\pm$0.082 \\
\cmidrule{2-8}          &       & MAE   & 0.967$\pm$0.067 & 1.014$\pm$0.056 & 0.976$\pm$0.082 & 0.953$\pm$0.039 & 0.959$\pm$0.059 \\
          & \multicolumn{1}{l}{LLMP Omit (No Text)} & ACC   & 0.265$\pm$0.112 & 0.427$\pm$0.048 & 0.475$\pm$0.028 & 0.440$\pm$0.077 & 0.470$\pm$0.035 \\
          &       & NLL   & 4.010$\pm$0.222 & 3.691$\pm$0.315 & 3.521$\pm$0.321 & 3.420$\pm$0.223 & 3.307$\pm$0.209 \\
\cmidrule{2-8}          &       & MAE   & 0.832$\pm$0.032 & 0.877$\pm$0.023 & 0.787$\pm$0.072 & 0.791$\pm$0.046 & 0.804$\pm$0.075 \\
    30    & \multicolumn{1}{l}{LLMP Impute (Text)} & ACC   & 0.437$\pm$0.035 & 0.528$\pm$0.047 & 0.522$\pm$0.046 & 0.523$\pm$0.033 & 0.508$\pm$0.007 \\
          &       & NLL   & 2.656$\pm$0.071 & 2.552$\pm$0.040 & 2.509$\pm$0.077 & 2.493$\pm$0.037 & 2.542$\pm$0.020 \\
\cmidrule{2-8}          &       & MAE   & 0.804$\pm$0.034 & 0.860$\pm$0.033 & 0.805$\pm$0.058 & 0.790$\pm$0.037 & 0.750$\pm$0.074 \\
          & \multicolumn{1}{l}{LLMP Omit (Text)} & ACC   & 0.437$\pm$0.027 & 0.505$\pm$0.058 & 0.510$\pm$0.062 & 0.525$\pm$0.035 & 0.520$\pm$0.035 \\
          &       & NLL   & 2.730$\pm$0.042 & 2.581$\pm$0.015 & 2.569$\pm$0.011 & 2.513$\pm$0.029 & 2.486$\pm$0.016 \\
\cmidrule{2-8}          &       & MAE   & 0.998$\pm$0.174 & 0.734$\pm$0.073 & 0.760$\pm$0.067 & 0.763$\pm$0.099 & 0.767$\pm$0.037 \\
          & \multicolumn{1}{l}{TabPFN} & ACC   & 0.292$\pm$0.070 & 0.425$\pm$0.030 & 0.440$\pm$0.024 & 0.475$\pm$0.016 & 0.462$\pm$0.019 \\
          &       & NLL   & 3.315$\pm$0.255 & 2.846$\pm$0.191 & 2.700$\pm$0.142 & 2.622$\pm$0.097 & 2.553$\pm$0.072 \\
    \midrule
          &       & MAE   & 0.966$\pm$0.040 & 0.976$\pm$0.042 & 0.956$\pm$0.077 & 0.931$\pm$0.026 & 0.928$\pm$0.077 \\
          & \multicolumn{1}{l}{LLMP Impute (No Text)} & ACC   & 0.284$\pm$0.069 & 0.438$\pm$0.028 & 0.468$\pm$0.021 & 0.462$\pm$0.047 & 0.480$\pm$0.018 \\
          &       & NLL   & 3.475$\pm$0.311 & 2.948$\pm$0.019 & 2.782$\pm$0.110 & 2.750$\pm$0.055 & 2.718$\pm$0.134 \\
\cmidrule{2-8}          &       & MAE   & 0.974$\pm$0.020 & 1.050$\pm$0.050 & 1.012$\pm$0.058 & 1.024$\pm$0.015 & 1.000$\pm$0.056 \\
          & \multicolumn{1}{l}{LLMP Omit (No Text)} & ACC   & 0.245$\pm$0.076 & 0.392$\pm$0.065 & 0.448$\pm$0.016 & 0.450$\pm$0.068 & 0.465$\pm$0.012 \\
          &       & NLL   & 4.229$\pm$0.595 & 3.777$\pm$0.251 & 3.656$\pm$0.244 & 3.563$\pm$0.161 & 3.539$\pm$0.193 \\
\cmidrule{2-8}          &       & MAE   & 0.876$\pm$0.020 & 0.945$\pm$0.036 & 0.827$\pm$0.063 & 0.861$\pm$0.041 & 0.862$\pm$0.074 \\
    40    & \multicolumn{1}{l}{LLMP Impute (Text)} & ACC   & 0.440$\pm$0.017 & 0.512$\pm$0.033 & 0.525$\pm$0.020 & 0.523$\pm$0.016 & 0.507$\pm$0.007 \\
          &       & NLL   & 2.763$\pm$0.034 & 2.635$\pm$0.003 & 2.536$\pm$0.029 & 2.552$\pm$0.029 & 2.600$\pm$0.041 \\
\cmidrule{2-8}          &       & MAE   & 0.865$\pm$0.017 & 0.930$\pm$0.028 & 0.846$\pm$0.028 & 0.838$\pm$0.013 & 0.803$\pm$0.068 \\
          & \multicolumn{1}{l}{LLMP Omit (Text)} & ACC   & 0.435$\pm$0.035 & 0.483$\pm$0.021 & 0.502$\pm$0.026 & 0.530$\pm$0.049 & 0.533$\pm$0.037 \\
          &       & NLL   & 2.836$\pm$0.018 & 2.678$\pm$0.005 & 2.628$\pm$0.016 & 2.596$\pm$0.032 & 2.551$\pm$0.048 \\
\cmidrule{2-8}          &       & MAE   & 1.080$\pm$0.224 & 0.823$\pm$0.117 & 0.791$\pm$0.042 & 0.809$\pm$0.064 & 0.821$\pm$0.035 \\
          & \multicolumn{1}{l}{TabPFN} & ACC   & 0.268$\pm$0.102 & 0.423$\pm$0.045 & 0.438$\pm$0.019 & 0.490$\pm$0.014 & 0.457$\pm$0.024 \\
          &       & NLL   & 3.351$\pm$0.217 & 2.926$\pm$0.183 & 2.724$\pm$0.091 & 2.660$\pm$0.087 & 2.616$\pm$0.076 \\
    \bottomrule
    \end{tabular}%
    \end{adjustbox}
  \end{sc}
  \end{tiny}
  \vskip -0.1in
\end{table*}%

% Table generated by Excel2LaTeX from sheet 'missing classification'
\begin{table*}[htbp]
  \centering
  \caption{\textbf{Missing Data Handling on Cars}. AUC for JoLT using the gemma-2-27B LLM and TabPFN as a function of shots and percentage of data missing completely-at-random (MCAR) on the Car classification dataset \citep{car_evaluation_19}. 200 test examples were used. Values are the mean and 95$\%$ confidence interval (CI) over 5 seeds that affect the training and test shot selection and the missing pattern.}
  \label{tab:car_missing}%
  \vskip 0.1in
  \begin{small}
  \begin{sc}
  \begin{adjustbox}{max width=\textwidth}
    \begin{tabular}{clccccc}
    \toprule
          &       & \multicolumn{5}{c}{\textbf{Shot}} \\
\cmidrule{3-7}    \textbf{Missing \%} & \textbf{Method} & \textbf{10} & \textbf{20} & \textbf{30} & \textbf{40} & \textbf{50} \\
    \midrule
          & JoLT Impute (No Text) & 0.710$\pm$0.083 & 0.791$\pm$0.026 & 0.818$\pm$0.023 & 0.856$\pm$0.029 & 0.862$\pm$0.002 \\
          & JoLT Omit (No Text) & 0.685$\pm$0.082 & 0.778$\pm$0.033 & 0.808$\pm$0.003 & 0.844$\pm$0.028 & 0.827$\pm$0.029 \\
    10    & JoLT Impute (Text) & 0.736$\pm$0.061 & 0.813$\pm$0.010 & 0.819$\pm$0.012 & 0.851$\pm$0.022 & 0.856$\pm$0.007 \\
          & JoLT Omit (Text) & 0.755$\pm$0.030 & 0.822$\pm$0.046 & 0.841$\pm$0.027 & 0.877$\pm$0.038 & 0.876$\pm$0.016 \\
          & TabPFN & 0.737$\pm$0.049 & 0.840$\pm$0.012 & 0.877$\pm$0.021 & 0.895$\pm$0.011 & 0.892$\pm$0.004 \\
    \midrule
          & JoLT Impute (No Text) & 0.632$\pm$0.061 & 0.722$\pm$0.077 & 0.720$\pm$0.068 & 0.791$\pm$0.019 & 0.793$\pm$0.034 \\
          & JoLT Omit (No Text) & 0.667$\pm$0.039 & 0.718$\pm$0.067 & 0.742$\pm$0.017 & 0.778$\pm$0.042 & 0.791$\pm$0.060 \\
    20    & JoLT Impute (Text) & 0.678$\pm$0.042 & 0.787$\pm$0.037 & 0.779$\pm$0.028 & 0.824$\pm$0.018 & 0.819$\pm$0.022 \\
          & JoLT Omit (Text) & 0.714$\pm$0.034 & 0.786$\pm$0.039 & 0.799$\pm$0.028 & 0.844$\pm$0.037 & 0.834$\pm$0.038 \\
          & TabPFN & 0.679$\pm$0.024 & 0.811$\pm$0.038 & 0.808$\pm$0.018 & 0.824$\pm$0.029 & 0.804$\pm$0.048 \\
    \midrule
          & JoLT Impute (No Text) & 0.589$\pm$0.046 & 0.643$\pm$0.053 & 0.665$\pm$0.017 & 0.723$\pm$0.024 & 0.712$\pm$0.014 \\
          & JoLT Omit (No Text) & 0.647$\pm$0.023 & 0.659$\pm$0.089 & 0.676$\pm$0.068 & 0.696$\pm$0.054 & 0.715$\pm$0.090 \\
    30    & JoLT Impute (Text) & 0.614$\pm$0.058 & 0.722$\pm$0.037 & 0.716$\pm$0.015 & 0.749$\pm$0.017 & 0.745$\pm$0.030 \\
          & JoLT Omit (Text) & 0.696$\pm$0.040 & 0.768$\pm$0.051 & 0.756$\pm$0.026 & 0.793$\pm$0.040 & 0.792$\pm$0.032 \\
          & TabPFN & 0.687$\pm$0.034 & 0.774$\pm$0.035 & 0.769$\pm$0.037 & 0.794$\pm$0.024 & 0.783$\pm$0.036 \\
    \midrule
          & JoLT Impute (No Text) & 0.613$\pm$0.048 & 0.692$\pm$0.048 & 0.692$\pm$0.086 & 0.706$\pm$0.032 & 0.718$\pm$0.044 \\
          & JoLT Omit (No Text) & 0.598$\pm$0.055 & 0.619$\pm$0.048 & 0.633$\pm$0.070 & 0.660$\pm$0.081 & 0.655$\pm$0.055 \\
    40    & JoLT Impute (Text) & 0.670$\pm$0.039 & 0.737$\pm$0.029 & 0.733$\pm$0.038 & 0.763$\pm$0.035 & 0.758$\pm$0.030 \\
          & JoLT Omit (Text) & 0.648$\pm$0.036 & 0.729$\pm$0.052 & 0.761$\pm$0.032 & 0.780$\pm$0.032 & 0.770$\pm$0.041 \\
          & TabPFN & 0.617$\pm$0.105 & 0.735$\pm$0.028 & 0.646$\pm$0.077 & 0.721$\pm$0.056 & 0.710$\pm$0.021 \\
    \bottomrule
    \end{tabular}%
    \end{adjustbox}
  \end{sc}
  \end{small}
  \vskip -0.1in
\end{table*}%

% Table generated by Excel2LaTeX from sheet 'Impute'
\begin{table*}[h]
  \centering
  \caption{\textbf{Imputation}: MAE as a function of $\%$ of missing data (MCAR) on the Paris 2024 Olympic Medals dataset for JoLT that uses the gemma-2-27B and four competive methods. Values are the mean and 95$\%$ confidence interval (CI) over 3 seeds that affect the training and test shot selection and the missing pattern.}
  \label{tab:imputation}%
  \vskip 0.1in
  \begin{small}
  \begin{sc}
  \begin{adjustbox}{max width=\textwidth}
    \begin{tabular}{lcccc}
    \toprule
          & \multicolumn{4}{c}{\textbf{Missing \%}} \\
\cmidrule{2-5}    \textbf{Method} & \textbf{10} & \textbf{20} & \textbf{30} & \textbf{40} \\
    \midrule
    Mean  & 3.398$\pm$1.190 & 3.523$\pm$0.784 & 3.835$\pm$0.762 & 3.796$\pm$0.540 \\
    kNN   & 1.728$\pm$0.505 & 1.833$\pm$0.226 & 2.291$\pm$0.636 & 2.279$\pm$0.604 \\
    Iterative & 1.544$\pm$0.026 & 1.844$\pm$0.161 & 2.033$\pm$0.515 & 2.400$\pm$0.492 \\
    MICE  & 1.679$\pm$0.436 & 2.138$\pm$0.579 & 2.254$\pm$0.681 & 2.523$\pm$0.867 \\
    JoLT  & 1.139$\pm$0.253 & 1.590$\pm$0.306 & 1.465$\pm$0.044 & 1.543$\pm$0.337 \\
    \bottomrule
    \end{tabular}%
    \end{adjustbox}
    \end{sc}
    \end{small}
\end{table*}%

%%%%%%%%%%%%%%%%%%%%%%%%%%%%%%%%%%%%%%%%%%%%%%%%%%%%%%%%%%%%%%%%%%%%%%%%%%%%%%%
%%%%%%%%%%%%%%%%%%%%%%%%%%%%%%%%%%%%%%%%%%%%%%%%%%%%%%%%%%%%%%%%%%%%%%%%%%%%%%%

\end{document}